\PassOptionsToPackage{unicode}{hyperref}
\PassOptionsToPackage{hyphens}{url}
\documentclass[
  11pt]{article}
\usepackage{xcolor}
\usepackage[margin=1in]{geometry}
\usepackage{amsmath,amssymb}
\usepackage{iftex}
\ifPDFTeX
  \usepackage[T1]{fontenc}
  \usepackage[utf8]{inputenc}
  \usepackage{textcomp} 
\else 
  \usepackage{unicode-math} 
  \defaultfontfeatures{Scale=MatchLowercase}
  \defaultfontfeatures[\rmfamily]{Ligatures=TeX,Scale=1}
\fi
\usepackage{lmodern}
\ifPDFTeX\else
\fi
\IfFileExists{upquote.sty}{\usepackage{upquote}}{}
\IfFileExists{microtype.sty}{
  \usepackage[]{microtype}
  \UseMicrotypeSet[protrusion]{basicmath} 
}{}
\makeatletter
\@ifundefined{KOMAClassName}{
  \IfFileExists{parskip.sty}{%
    \usepackage{parskip}
  }{
    \setlength{\parindent}{0pt}
    \setlength{\parskip}{6pt plus 2pt minus 1pt}}
}{
  \KOMAoptions{parskip=half}}
\makeatother
\providecommand{\tightlist}{%
  \setlength{\itemsep}{0pt}\setlength{\parskip}{0pt}}
\usepackage[vvarbb, amsthm, cochineal, frenchmath, smallerops]{newtx}
\usepackage{lettrine}
\usepackage[utf8]{inputenc}
\usepackage{newunicodechar}
\usepackage{mathpartir}

\newunicodechar{λ}{\ensuremath{\lambda}}
\newunicodechar{Π}{\ensuremath{\Pi}}
\newunicodechar{Σ}{\ensuremath{\Sigma}}
\newunicodechar{∘}{\ensuremath{\circ}}
\newunicodechar{∀}{\ensuremath{\forall}}
\newunicodechar{→}{\ensuremath{\to}}
\newunicodechar{≡}{\ensuremath{\equiv}}
\newunicodechar{ℓ}{\ensuremath{\ell}}
\newunicodechar{₀}{\ensuremath{_0}}
\newunicodechar{₁}{\ensuremath{_1}}
\newunicodechar{₂}{\ensuremath{_2}}
\newunicodechar{₃}{\ensuremath{_3}}
\newunicodechar{₄}{\ensuremath{_4}}
\newunicodechar{₅}{\ensuremath{_5}}
\newunicodechar{⊔}{\ensuremath{\sqcup}}
\newunicodechar{⊤}{\ensuremath{\top}}
\newunicodechar{≐}{\ensuremath{\doteq}}
\newunicodechar{κ}{\ensuremath{\kappa}}
\newunicodechar{Δ}{\ensuremath{\Delta}}
\newunicodechar{Λ}{\ensuremath{\Lambda}}
\newunicodechar{δ}{\ensuremath{\delta}}
\newunicodechar{≤}{\ensuremath{\leq}}
\newunicodechar{∈}{\ensuremath{\in}}
\newunicodechar{∩}{\ensuremath{\cap}}
\newunicodechar{¬}{\ensuremath{\neg}}
\newunicodechar{∧}{\ensuremath{\wedge}}
\newunicodechar{∨}{\ensuremath{\vee}}
\newunicodechar{β}{\ensuremath{\beta}}

\theoremstyle{definition}

\theoremstyle{remark}

\newunicodechar{⊥}{\ensuremath{\bot}}
\newunicodechar{⊕}{\ensuremath{\oplus}}
\newunicodechar{⊙}{\ensuremath{\odot}}
\newunicodechar{⇒}{\ensuremath{\Rightarrow}}
\newunicodechar{×}{\ensuremath{\times}}
\newunicodechar{∣}{\ensuremath{\mid}}
\newunicodechar{⊗}{\ensuremath{\otimes}}

\usepackage{booktabs}
\usepackage{dashrule}
\usepackage{fourier-orns}
\usepackage{quiver}

\usepackage{fancyhdr}
\usepackage[explicit]{titlesec}

\titleformat{\section}{\normalfont\Large}{\aldine \enspace \thesection}{1ex}{#1}
\titlespacing{\section}{0pt}{1ex}{1em}

\titleformat{\subsection}{\normalfont\large}{\aldinesmall \enspace \thesubsection}{1ex}{#1}
\titlespacing{\subsection}{0pt}{1ex}{1em}

\titleformat{\subsubsection}[runin]{\normalfont\normalsize\bfseries}{$\S$ \enspace \thesubsubsection}{
    \ifstrempty{#1}{0pt}{1em}}{
    \ifstrempty{#1}{#1}{\guillemets{#1}}}
\titlespacing{\subsubsection}{0pt}{1ex}{1em}

\titleformat{\paragraph}[runin]{\normalfont\normalsize\bfseries}{}{0pt}{\guillemets{#1}}
\titlespacing{\paragraph}{0pt}{1ex}{1em}

\titleformat{\subparagraph}[runin]{\normalfont\normalsize\bfseries}{\lefthand}{1em}{#1}
\titlespacing{\subparagraph}{0pt}{1ex}{1em}

\renewcommand{\thesection}{\arabic{section}}
\renewcommand{\thesubsection}{\arabic{section}.\arabic{subsection}}
\renewcommand{\thesubsubsection}{\arabic{section}.\arabic{subsection}.\arabic{subsubsection}}

\newcommand{\secend}{\noindent\centerline{\Large \decotwo}\\}

\newcommand{\guillemets}[1]{\guillemetleft #1\guillemetright}

\usepackage{lipsum}
\usepackage{blindtext}
\usepackage{bookmark}
\IfFileExists{xurl.sty}{\usepackage{xurl}}{} 
\hypersetup{
  pdftitle={Look buddy{,} it's transient{,} shifting like water},
  pdfauthor={Corinthia Aberlé, Emma Quinn, Artun Saday},
  hidelinks,
  pdfcreator={LaTeX via pandoc}}

\title{Look buddy, it's transient, shifting like water}
\usepackage{etoolbox}
\makeatletter
\providecommand{\subtitle}[1]{
  \apptocmd{\@title}{\par {\large #1 \par}}{}{}
}
\makeatother
\subtitle{Learning Dimensions for Punctuated Equilibria}
\author{Corinthia Aberlé\\ \small Carnegie Mellon University\\ \small\texttt{caberle@andrew.cmu.edu} \and Emma Quinn\\ \small Carnegie Mellon University\\ \small\texttt{gquinn@andrew.cmu.edu} \and Artun Saday\\ \small Carnegie Mellon University\\ \small\texttt{asaday@andrew.cmu.edu}}
\date{\today}

\begin{document}
\maketitle
\begin{abstract}
VC dimension counts the questions a hypothesis class can answer in every
possible way when they are posed all at once, and governs learning when
examples are drawn i.i.d. from a fixed distribution; Littlestone
dimension counts them when each question may depend on the answers to
the previous ones, and governs learning when an adversary chooses every
example. We introduce the \emph{epoch dimension} \(\mathrm{ED}_n(H)\),
which measures questions posed in at most \(n\) epochs, all at once
within an epoch and adaptively across epochs, with every answer given by
a single hypothesis. It equals the VC dimension for \(n=1\) and the
Littlestone dimension once \(n\ge\mathrm{LD}(H)\), and in between it can
lie strictly between the two. We compute it for thresholds, unions of
intervals, and cubes, prove that it is additive over products, and show
that for half-spaces in \(\mathbb{R}^d\) it is \(dn+1\). We then show
that it characterizes learning when a fixed concept labels examples
whose distribution shifts at up to \(n-1\) unannounced times: every
learner can be forced to make \(\Omega(\mathrm{ED}_n(H))\) expected
mistakes, and a layered version of Littlestone's standard optimal
algorithm, aggregated over possible restart times by fixed share, makes
\(O(\mathrm{ED}_n(H)\log T)\) expected mistakes in \(T\) rounds, without
knowing \(n\).
\end{abstract}

\makeatletter
\newcommand{\@envopt}[2]{\if\relax\detokenize{#2}\relax\expandafter\@firstoftwo\else\expandafter\@secondoftwo\fi{\begin{#1}}{\begin{#1}[{#2}]}}
\newcommand{\Thm}[1]{\@envopt{theorem}{#1}}\newcommand{\EndThm}{\end{theorem}}
\newcommand{\Prop}[1]{\@envopt{proposition}{#1}}\newcommand{\EndProp}{\end{proposition}}
\newcommand{\Lem}[1]{\@envopt{lemma}{#1}}\newcommand{\EndLem}{\end{lemma}}
\newcommand{\Cor}[1]{\@envopt{corollary}{#1}}\newcommand{\EndCor}{\end{corollary}}
\newcommand{\Defn}[1]{\@envopt{definition}{#1}}\newcommand{\EndDefn}{\end{definition}}
\newcommand{\Ex}[1]{\@envopt{example}{#1}}\newcommand{\EndEx}{\end{example}}
\newcommand{\Rem}[1]{\@envopt{remark}{#1}}\newcommand{\EndRem}{\end{remark}}
\newcommand{\Pf}[1]{\@envopt{proof}{#1}}\newcommand{\EndPf}{\end{proof}}
\makeatother
\providecommand{\VC}{\mathrm{VC}}
\providecommand{\Ldim}{\mathrm{LD}}
\providecommand{\ED}{\mathrm{ED}}
\providecommand{\speaker}[1]{\textsc{#1}.\enspace}
\providecommand{\E}{\mathop{\mathbb{E}}\nolimits}
\providecommand{\Opt}{\mathrm{Opt}}

\section{Introduction}\label{sec:intro}

\begin{flushright}
\begin{minipage}{0.62\textwidth}
\small\raggedright
\hangindent=1.5em\hangafter=1
\speaker{Me} Is the natural state of the soul quiet or chaos?\par
\smallskip
\hangindent=1.5em\hangafter=1
\speaker{Taco Bell cashier} Look buddy, it's transient, shifting like water\par
\medskip
\raggedleft ---surf ghost (@moutheaters), Twitter 2017\par
\end{minipage}
\end{flushright}
\medskip

Fix a set \(Q\) of points and a class \(H\subseteq\{0,1\}^Q\). The VC
dimension \(\VC(H)\) is the largest size of a set of points all of whose
labelings are realized by hypotheses in \(H\) {[}VC71{]}. The
Littlestone dimension \(\Ldim(H)\) is the largest depth of a complete
binary tree of points every root-to-leaf path of which, read as a
sequence of labels, is realized by a hypothesis {[}Lit88{]}. Each
governs learning in one state of the world. In times of \emph{quiet},
the examples are drawn i.i.d. from a fixed distribution, and the VC
dimension governs learning {[}VC71; HLW94{]}. In times of \emph{chaos},
an adversary chooses each example after seeing the answers to the
previous ones, and the Littlestone dimension is the optimal mistake
bound {[}Lit88{]}. Both dimensions count questions ``what is the label
of \(x\)?'' whose every combination of answers is realized by a single
hypothesis. They differ in whether the questions must be chosen before
any answer is known, or may each depend on the answers so far (Figure
\ref{fig:quietchaos}).

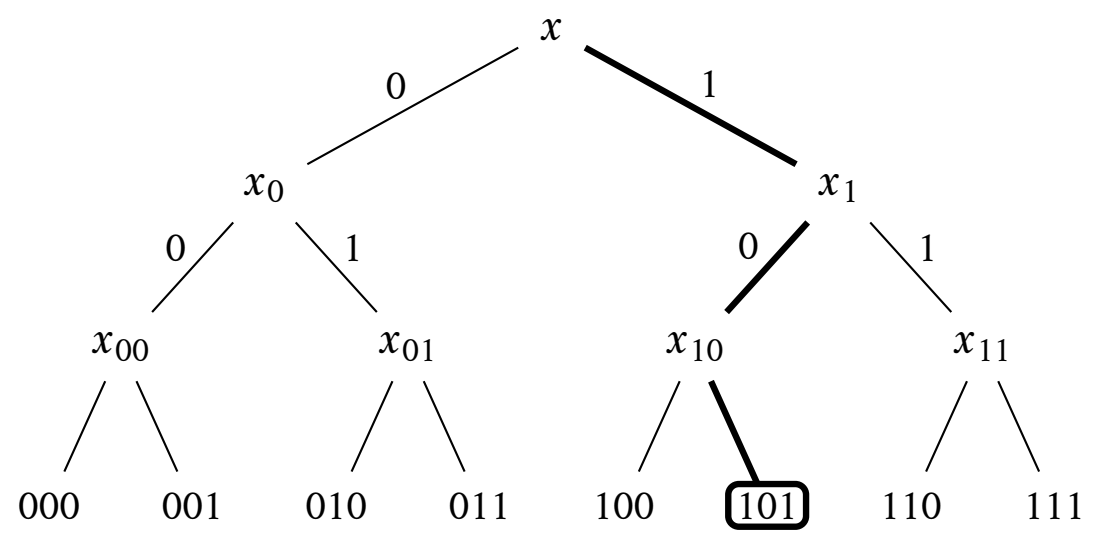
\begin{figure}[ht]
\centering
\begin{tikzpicture}[every node/.style={font=\footnotesize}]
\begin{scope}
\foreach \c/\n in {0.5/1,1.0/2,1.5/3} {\node at (\c,0.5) {$x_\n$};}
\draw (0.25,0.25) -- (1.75,0.25);
\foreach \r/\a/\b/\c in {0/0/0/0,1/0/0/1,2/0/1/0,3/0/1/1,4/1/0/0,5/1/0/1,6/1/1/0,7/1/1/1} {
  \node[anchor=east,font=\scriptsize] at (0.15,-0.44*\r) {$h_{\a\b\c}$};
  \node at (0.5,-0.44*\r) {$\a$};
  \node at (1.0,-0.44*\r) {$\b$};
  \node at (1.5,-0.44*\r) {$\c$};
}
\draw[thick,rounded corners=2pt] (0.3,-0.44*5+0.19) rectangle (1.75,-0.44*5-0.19);
\node at (0.7,-3.95) {(a) quiet: a shattered set};
\end{scope}
\begin{scope}[xshift=7.4cm,level distance=10.5mm,
  level 1/.style={sibling distance=38mm},
  level 2/.style={sibling distance=19mm},
  level 3/.style={sibling distance=9.5mm},
  lab/.style={font=\scriptsize,inner sep=1pt},
  hl/.style={edge from parent/.style={draw,very thick}}]
\node {$x$}
  child {node {$x_0$}
    child {node {$x_{00}$}
      child {node[font=\scriptsize] {$000$}}
      child {node[font=\scriptsize] {$001$}}
      edge from parent node[lab,above left] {$0$}}
    child {node {$x_{01}$}
      child {node[font=\scriptsize] {$010$}}
      child {node[font=\scriptsize] {$011$}}
      edge from parent node[lab,above right] {$1$}}
    edge from parent node[lab,above left] {$0$}}
  child[hl] {node {$x_1$}
    child[hl] {node {$x_{10}$}
      child[edge from parent/.style={draw,thin}] {node[font=\scriptsize] {$100$}}
      child[hl] {node[font=\scriptsize,draw,rounded corners=2pt,inner sep=1.5pt] {$101$}}
      edge from parent node[lab,above left] {$0$}}
    child[edge from parent/.style={draw,thin}] {node {$x_{11}$}
      child {node[font=\scriptsize] {$110$}}
      child {node[font=\scriptsize] {$111$}}
      edge from parent node[lab,above right] {$1$}}
    edge from parent node[lab,above right] {$1$}};
\node at (0,-3.95) {(b) chaos: a shattered Littlestone tree};
\end{scope}
\end{tikzpicture}
\caption{Quiet and chaos. (a) A fixed distribution chooses its questions before any answer is known, so in effect it poses them all at once. (b) An adversary poses the questions one at a time, and the subscript of each point records the answers so far, so which point it asks next depends on them.}
\label{fig:quietchaos}
\end{figure}

Is the natural state of the world quiet or chaos? Eldredge and Gould
proposed an answer for the history of life {[}EG72; GE77{]}. Against
\emph{phyletic gradualism}, the view that species evolve slowly and
steadily, they read the fossil record as showing \emph{punctuated
equilibria}: species persist almost unchanged through long stretches of
geological time, and new species arise in brief episodes of (relatively)
rapid change. On this view the world is transient, shifting like
water---long stretches of quiet broken by sudden shifts.

We adapt Eldredge and Gould's concept to form a bridge between the VC
and Littlestone regimes. Suppose a single unknown concept \(c\in H\),
the \emph{law}, labels every example, but the distribution of the
examples may change abruptly at up to \(n-1\) times chosen by an
adversary, dividing time into at most \(n\) epochs, and within each
epoch the examples are i.i.d. Within an epoch the world is quiet. Chaos
enters only at the punctuations. The epoch dimension \(\ED_n(H)\)
measures how much the adversary can extract. It counts questions posed
in at most \(n\) epochs, all at once within an epoch and adaptively
across epochs, and it runs from the VC dimension for \(n=1\) to the
Littlestone dimension once \(n\ge\Ldim(H)\).

Our main result, Theorem \ref{thm:main}, is that it characterizes this
setting up to a logarithmic factor: every learner, even one told where
the epochs begin, can be forced to make \(\frac{1-1/e}{2}\ED_n(H)\)
expected mistakes (Section \ref{sec:lower}), and a layered version of
Littlestone's standard optimal algorithm, aggregated over possible
restart times by fixed share, makes at most \(12(1+\ln T)\ED_n(H)\) in
\(T\) rounds without being told, and without knowing \(n\) (Section
\ref{sec:upper}). The upper bound rests on a trade-off between present
and future uncertainty (Lemma \ref{lem:highvalue}): the labelings of an
epoch's points that leave much epoch dimension for the later epochs form
a class of small VC dimension. The layered algorithm predicts as
Littlestone's does when one label leaves more for the future than the
other, and as the one-inclusion graph predictor does on these simple
classes when the two tie; so each epoch costs, up to a logarithmic
factor, only as much as it uses up of the epoch dimension (Figure
\ref{fig:pe1}).

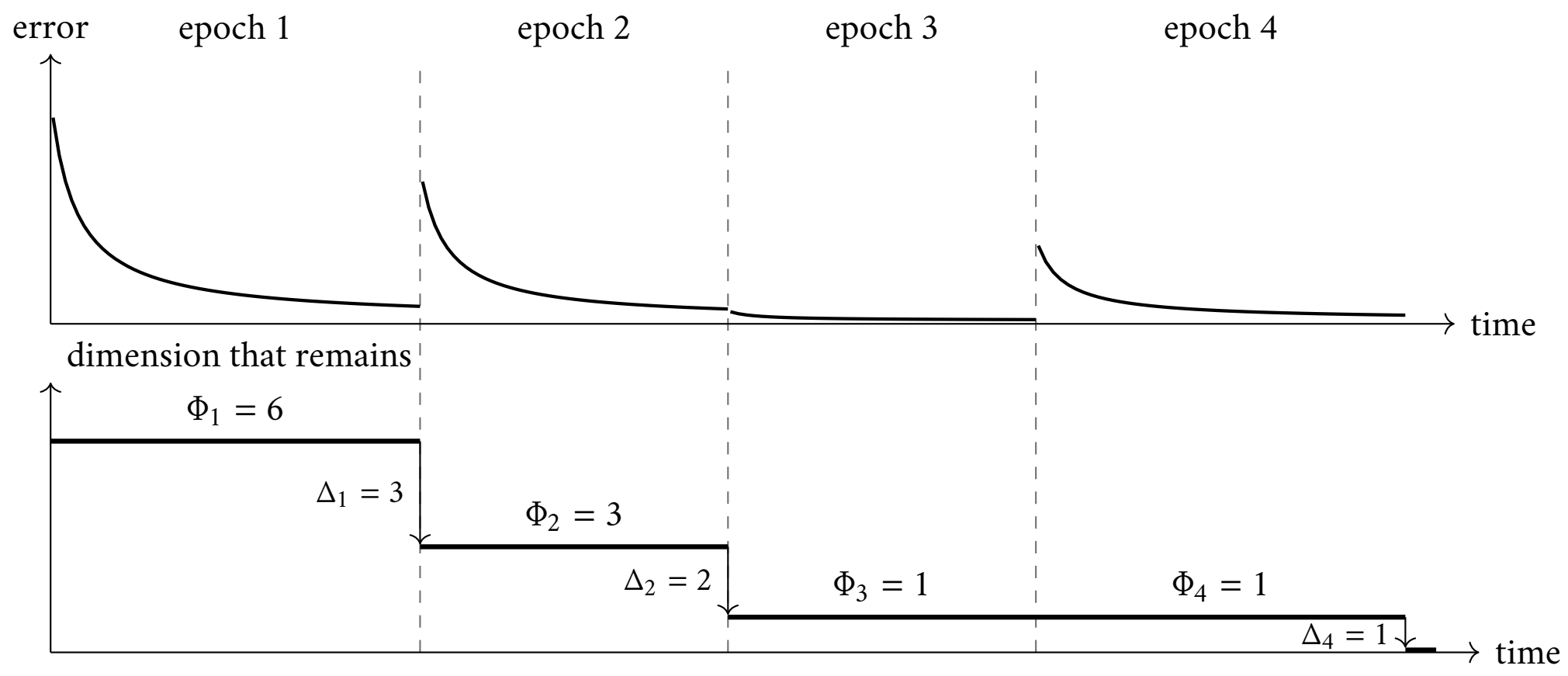
\begin{figure}[ht]
\centering
\begin{tikzpicture}[x=1.05cm,y=1cm,every node/.style={font=\footnotesize}]
\foreach \t in {3,5.5,8} {\draw[dashed,black!55] (\t,0) -- (\t,5.0);}
\foreach \c/\n in {1.5/1,4.25/2,6.75/3,9.5/4} {\node at (\c,5.3) {epoch \n};}
\draw[->] (0,2.8) -- (11.4,2.8) node[right] {time};
\draw[->] (0,2.8) -- (0,5.1) node[above] {error};
\draw[thick,domain=0.02:3,samples=60] plot (\x,{2.83+1.9/(1+5*\x)});
\draw[thick,domain=3.02:5.5,samples=50] plot (\x,{2.83+1.3/(1+5*(\x-3))});
\draw[thick,domain=5.52:8,samples=40] plot (\x,{2.83+0.08/(1+5*(\x-5.5))});
\draw[thick,domain=8.02:11,samples=50] plot (\x,{2.83+0.7/(1+5*(\x-8))});
\draw[->] (0,0) -- (11.6,0) node[right] {time};
\draw[->] (0,0) -- (0,2.3) node[above right] {dimension that remains};
\draw[very thick] (0,1.8) -- (3,1.8) (3,0.9) -- (5.5,0.9) (5.5,0.3) -- (11,0.3);
\draw[very thick] (11,0.02) -- (11.25,0.02);
\draw[thin,->] (3,1.8) -- (3,0.92);
\draw[thin,->] (5.5,0.9) -- (5.5,0.32);
\draw[thin,->] (11,0.3) -- (11,0.04);
\node[above] at (1.5,1.8) {$\Phi_1=6$};
\node[above] at (4.25,0.9) {$\Phi_2=3$};
\node[above] at (6.75,0.3) {$\Phi_3=1$};
\node[above] at (9.5,0.3) {$\Phi_4=1$};
\node[left,font=\scriptsize] at (3,1.35) {$\Delta_1=3$};
\node[left,font=\scriptsize] at (5.5,0.6) {$\Delta_2=2$};
\node[left,font=\scriptsize] at (11,0.14) {$\Delta_4=1$};
\end{tikzpicture}
\caption{Punctuated equilibria under one law (schematic). At each punctuation (dashed) the adversary shifts the distribution from which samples are drawn. \emph{Top:} the learner's error spikes when the distribution shifts and decays again within the epoch. \emph{Bottom:} the epoch dimension of the hypotheses still consistent with the data, for the epochs that remain.}
\label{fig:pe1}
\end{figure}

\section{Epoch trees and the epoch dimension}\label{sec:def}

A \emph{labeling} of a tuple \((x_1,\dots,x_s)\) of points is a string
\(\mu\in\{0,1\}^s\); a hypothesis \(h\) \emph{realizes} it if
\(h(x_k)=\mu_k\) for all \(k\).

\Defn{Epoch tree}

\label{def:epochtree} Let \(n\ge1\). An \emph{epoch tree} of depth \(n\)
is a rooted tree with \(n\) levels in which every node \(v\) carries a
tuple \(A_v\) of points, possibly empty, and every node above the last
level has one child for each labeling of its tuple. A \emph{path}
chooses, at each node \(v\) it visits, a labeling \(\mu_v\) of \(A_v\),
and moves on to the corresponding child. The tree has \emph{\(N\)
points} if every path contains exactly \(N\) points, counted with
multiplicity. Figure \ref{fig:histpos} shows an example: the lengths of
epochs may depend on the answers, but the total number of points along a
path may not.

\EndDefn

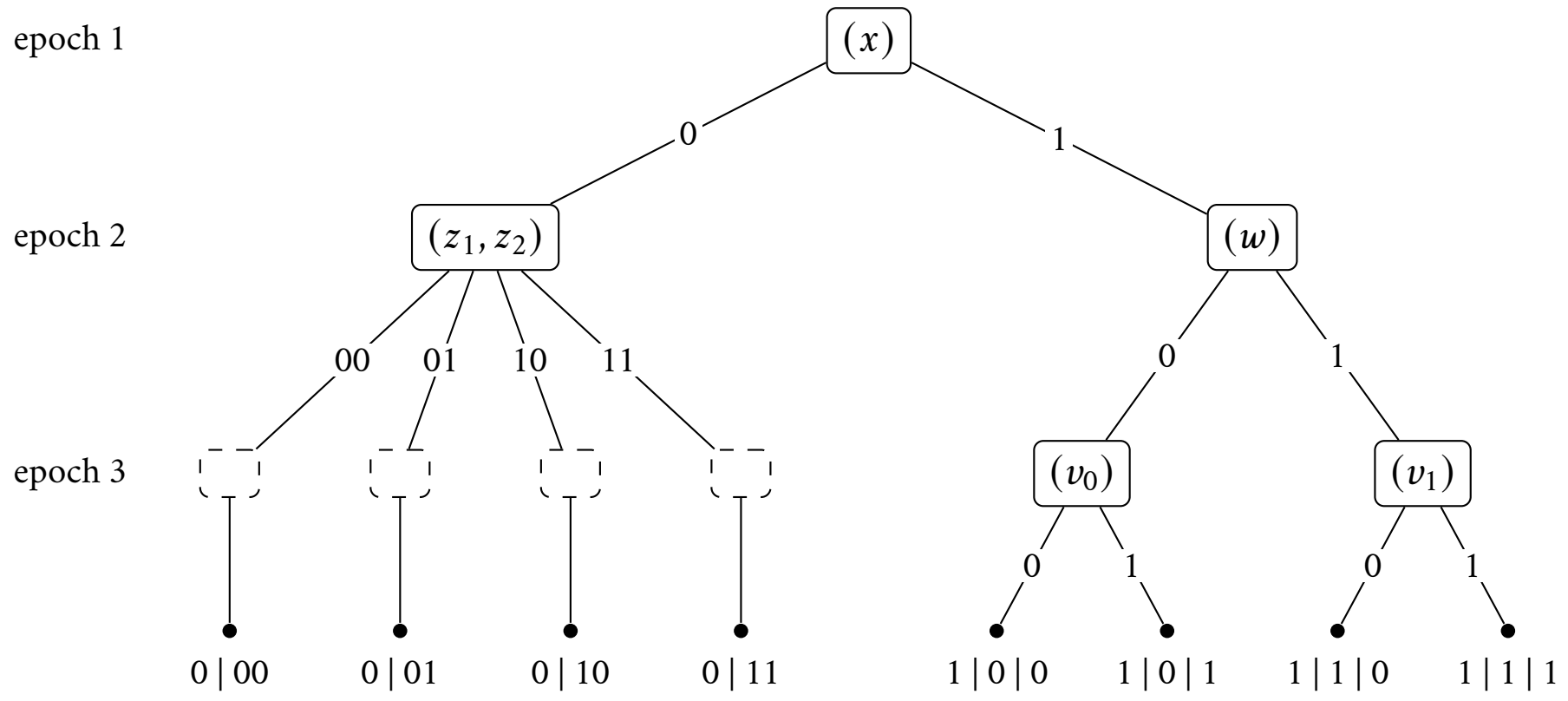
\begin{figure}[ht]
\centering
\begin{tikzpicture}[x=1.3cm,y=1cm,every node/.style={font=\footnotesize},
  blk/.style={draw,rounded corners=2pt,inner sep=3pt},
  emp/.style={draw,dashed,rounded corners=2pt,minimum width=4.5mm,minimum height=3.6mm,inner sep=0pt},
  leaf/.style={circle,fill,inner sep=1.1pt},
  lab/.style={font=\scriptsize,fill=white,inner sep=1pt}]
\node[font=\scriptsize,anchor=east] at (-0.5,0) {epoch 1};
\node[font=\scriptsize,anchor=east] at (-0.5,-1.5) {epoch 2};
\node[font=\scriptsize,anchor=east] at (-0.5,-3.3) {epoch 3};
\node[blk] (r) at (3.75,0) {$(x)$};
\node[blk] (z) at (1.5,-1.5) {$(z_1,z_2)$};
\node[blk] (w) at (6,-1.5) {$(w)$};
\draw (r) -- node[lab] {$0$} (z);
\draw (r) -- node[lab] {$1$} (w);
\foreach \idx/\lbl in {0/00,1/01,2/10,3/11} {
  \node[emp] (e\idx) at (\idx,-3.3) {};
  \node[leaf] (l\idx) at (\idx,-4.5) {};
  \draw (z) -- node[lab,pos=0.5] {$\lbl$} (e\idx);
  \draw (e\idx) -- (l\idx);
  \node[font=\scriptsize,below=2pt] at (l\idx) {$0\,|\,\lbl$};
}
\node[blk] (v0) at (5,-3.3) {$(v_0)$};
\node[blk] (v1) at (7,-3.3) {$(v_1)$};
\draw (w) -- node[lab] {$0$} (v0);
\draw (w) -- node[lab] {$1$} (v1);
\foreach \idx/\xx/\pp/\lbl in {4/4.5/0/0,5/5.5/0/1,6/6.5/1/0,7/7.5/1/1} {
  \node[leaf] (l\idx) at (\xx,-4.5) {};
  \draw (v\pp) -- node[lab] {$\lbl$} (l\idx);
  \node[font=\scriptsize,below=2pt] at (l\idx) {$1\,|\,\pp\,|\,\lbl$};
}
\end{tikzpicture}
\caption{An epoch tree of depth $3$ with $3$ points. The first epoch asks about $x$. After the answer $0$, the second epoch asks about $z_1$ and $z_2$ together, and the third epoch is empty (dashed); after the answer $1$, the second epoch asks about $w$ alone, and the third about $v_0$ or $v_1$, depending on the answer. So the lengths of the epochs depend on the answers, but every path meets exactly three points.}
\label{fig:histpos}
\end{figure}

\Defn{Epoch dimension}

\label{def:epochdim} An epoch tree is \emph{shattered by \(H\) under one
law} if for every path some single \(h\in H\) realizes the labeling
\(\mu_v\) of \(A_v\) at every node \(v\) the path visits. The
\emph{epoch dimension of depth \(n\)}, \(\ED_n(H)\), is the supremum of
the \(N\) such that some epoch tree of depth \(n\) with \(N\) points is
shattered by \(H\) under one law.

\EndDefn

Each level is an epoch: its tuple is a batch of questions posed
together, and which batch comes next depends on all the answers so far.
An epoch tree of depth \(1\) is a single tuple, so \(\ED_1(H)=\VC(H)\);
an epoch tree whose tuples are single points is a Littlestone tree.
Figure \ref{fig:shatter} shows the role of the single law.

\begin{figure}[ht]
\centering
\begin{tikzpicture}[x=1cm,y=1cm,every node/.style={font=\footnotesize},
  blk/.style={draw,rounded corners=2pt,inner sep=3pt},
  leaf/.style={circle,fill,inner sep=1.1pt},
  lab/.style={font=\scriptsize,fill=white,inner sep=1pt}]
\begin{scope}
\node[font=\small] at (-0.3,0.6) {(a)};
\node[blk] (r) at (2.25,0) {$\tfrac12$};
\node[blk] (a) at (0.75,-1.4) {$\tfrac34$};
\node[blk] (b) at (3.75,-1.4) {$\tfrac14$};
\draw (r) -- node[lab] {$0$} (a);
\draw (r) -- node[lab] {$1$} (b);
\foreach \idx/\xx/\pp/\lbl/\txt in {0/0/a/0/{(\tfrac34,\infty)},1/1.5/a/1/{(\tfrac12,\tfrac34]},2/3/b/0/{(\tfrac14,\tfrac12]},3/4.5/b/1/{(-\infty,\tfrac14]}} {
  \node[leaf] (l\idx) at (\xx,-2.7) {};
  \draw (\pp) -- node[lab] {$\lbl$} (l\idx);
  \node[font=\scriptsize,below=2pt] at (l\idx) {$\txt$};
}
\end{scope}
\begin{scope}[xshift=6.6cm]
\node[font=\small] at (-0.3,0.6) {(b)};
\node[blk] (r) at (2.25,0) {$\tfrac12$};
\node[blk] (a) at (0.75,-1.4) {$\tfrac14$};
\node[blk] (b) at (3.75,-1.4) {$\tfrac34$};
\draw (r) -- node[lab] {$0$} (a);
\draw (r) -- node[lab] {$1$} (b);
\foreach \idx/\xx/\pp/\lbl/\txt in {0/0/a/0/{(\tfrac12,\infty)},1/1.5/a/1/{\times},2/3/b/0/{\times},3/4.5/b/1/{(-\infty,\tfrac12]}} {
  \node[leaf] (l\idx) at (\xx,-2.7) {};
  \draw (\pp) -- node[lab] {$\lbl$} (l\idx);
  \node[font=\scriptsize,below=2pt] at (l\idx) {$\txt$};
}
\end{scope}
\end{tikzpicture}
\caption{Two epoch trees of depth $2$ with one point per epoch, for the thresholds $h_\theta(x)=\mathbf{1}[x\ge\theta]$. (a) Binary search. Each path is realized by every threshold $\theta$ in the interval below it, so the tree is shattered under one law. (b) Here the second question is one whose answer the first answer already settles. The crossed paths would need $\theta>\frac12$ and $\theta\le\frac14$, or $\theta\le\frac12$ and $\theta>\frac34$, so the tree is not shattered under one law. (If each epoch had a law of its own, it would be; see Remark~\ref{rem:perepoch}.)}
\label{fig:shatter}
\end{figure}
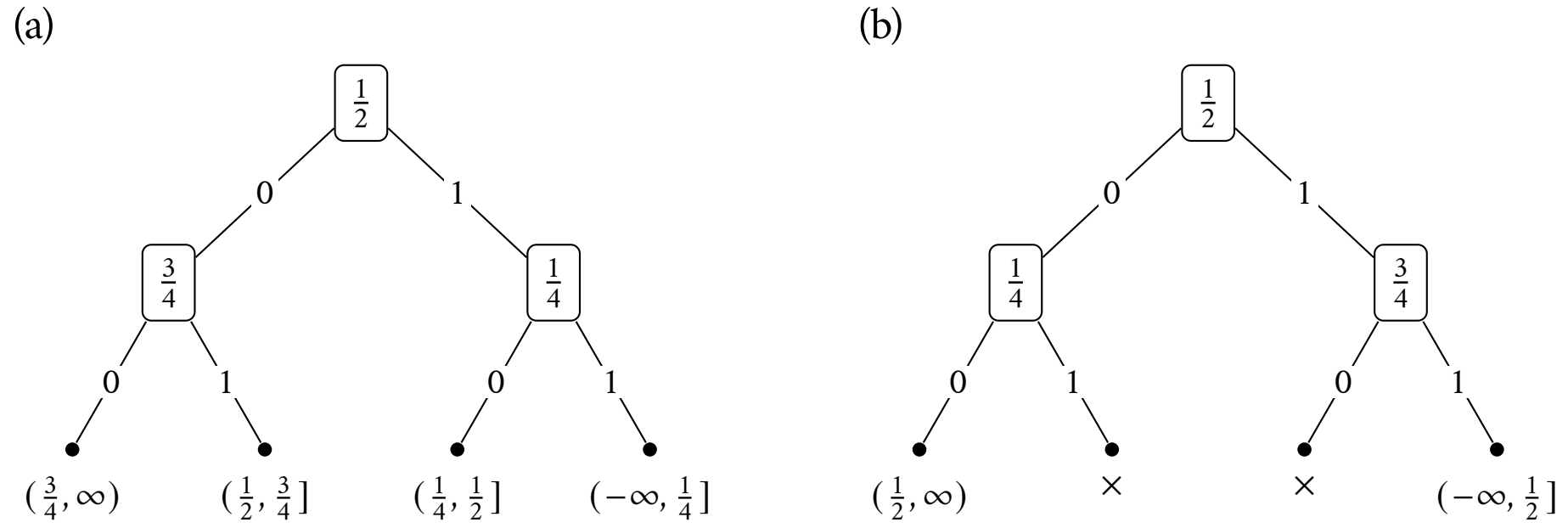

\Rem{A law per epoch}

\label{rem:perepoch} If each path need only be realized by a tuple of
hypotheses, one for each level, then a tree is shattered iff the tuple
at every node is shattered, and the corresponding dimension is simply
\(n\VC(H)\). The single law is what makes earlier answers constrain
later ones.

\EndRem

\Lem{Shattered epoch trees}

\label{lem:norepeat} In an epoch tree shattered by \(H\) under one law,
the tuple at every node consists of distinct points and is shattered by
\(H\), and no point occurs twice along a path.

\EndLem

\Pf{}

Every labeling of the tuple at a node \(v\) begins some path, which is
realized; so the tuple is shattered, and a repeated point could not
receive two labels. If a point \(a\) occurs at \(v\) and again at a node
\(w\) below \(v\), every hypothesis realizing a path through \(w\) gives
\(a\) the label it has along the way from \(v\) to \(w\), so the paths
through \(w\) that give \(a\) the other label at \(w\) are not realized.
\EndPf

\section{Basic properties}\label{sec:properties}

For a class \(V\subseteq\{0,1\}^Q\), a finite set \(A\subseteq Q\), and
a labeling \(\mu\) of \(A\), let
\(V_{A\to\mu}=\{h\in V:h\text{ realizes }\mu\text{ on }A\}\), the
hypotheses that remain once the labels of \(A\) are known, and write
\(V_{x\to b}\) for \(V_{\{x\}\to b}\). The \emph{version space} after
some labeled examples is the set of hypotheses of \(H\) consistent with
all of them.

\Prop{Recursion}

\label{prop:recursion} Let \(V\) be nonempty, and set \(\ED_0(V)=0\).
For every \(n\ge1\),
\[\ED_{n}(V)\;=\;\sup_{A ~ \text{shattered by} ~ V}\Big(|A|+\min_{\mu \in \{0,1\}^A}\ED_{n-1}(V_{A\to\mu})\Big),\]
Moreover, if \(V\) shatters an epoch tree of depth \(n\) with \(N\)
points, then for every \(N'\le N\) it shatters one with \(N'\) points.

\EndProp

\Pf{}

For the last claim, cut every path after its first \(N'\) points,
keeping an initial segment of the tuple where the path reaches its
\(N'\)-th point and emptying all later tuples; the cuts are consistent
across paths, and every new path extends to an old one. For the
recursion, if \(A\) is shattered and each \(V_{A\to\mu}\) shatters a
tree \(R_\mu\) of depth \(n-1\) with \(N'\) points, then the tree with
\(A\) at the root and \(R_\mu\) below the child for \(\mu\) is
shattered, since the witnesses of the paths of \(R_\mu\) lie in
\(V_{A\to\mu}\); by the last claim, \(N'\) may be any integer up to
\(\min_\mu\ED_{n-1}(V_{A\to\mu})\). Conversely, the root of a shattered
tree with \(N\) points carries a set \(A\) shattered by \(V\) (Lemma
\ref{lem:norepeat}), and below the child for \(\mu\) hangs a tree of
depth \(n-1\) with \(N-|A|\) points shattered by \(V_{A\to\mu}\). \EndPf

\Prop{Between VC and Littlestone dimension}

\label{prop:bounds} Let \(\VC(H)=d<\infty\). Then \(\ED_n(H)\) is
non-decreasing in \(n\) and in \(H\), and for every \(n\ge1\)
\[\max\big(d,\ \min(\Ldim(H),n)\big)\;\le\;\ED_n(H)\;\le\;\min\big(\Ldim(H),\ nd\big).\]
In particular \(\ED_1(H)=d\), and \(\ED_n(H)=\Ldim(H)\) once
\(n\ge\Ldim(H)\).

\EndProp

\Pf{}

Adding a level with an empty tuple at the top of an epoch tree gives
monotonicity in \(n\), and monotonicity in \(H\) is clear. A shattered
Littlestone tree of depth \(L=\min(\Ldim(H),n)\), padded with empty
levels, is a shattered epoch tree with \(L\) points. By Lemma
\ref{lem:norepeat}, every path meets \(n\) shattered tuples, so
\(N\le nd\). Finally, \emph{sequentializing} (Figure \ref{fig:seqtree}),
that is, replacing each tuple \((x_1,\dots,x_s)\) by a complete binary
tree of depth \(s\) whose nodes at depth \(k\) carry \(x_k\), and
hanging below the leaf for \(\mu\) the subtree for \(\mu\), turns a
shattered epoch tree with \(N\) points into a shattered Littlestone tree
of depth \(N\); so \(N\le\Ldim(H)\). \EndPf

\begin{figure}[ht]
\centering
\begin{tikzpicture}[every node/.style={font=\footnotesize},
  level distance=12mm,
  blk/.style={draw,rounded corners=2pt,inner sep=3pt},
  lab/.style={font=\scriptsize,fill=white,inner sep=1pt}]
\begin{scope}[level 1/.style={sibling distance=15mm}]
\node[blk] {$(x_1,x_2)$}
  child {node[blk] {$(z_{00})$} edge from parent node[lab] {$00$}}
  child {node[blk] {$(z_{01})$} edge from parent node[lab] {$01$}}
  child {node[blk] {$(z_{10})$} edge from parent node[lab] {$10$}}
  child {node[blk] {$(z_{11})$} edge from parent node[lab] {$11$}};
\node at (0,-2.4) {(a) an epoch tree of depth $2$};
\end{scope}
\draw[->,thick] (3.3,-0.7) -- node[above] {sequentialize} (5.0,-0.7);
\begin{scope}[xshift=8.6cm,level distance=10mm,
  level 1/.style={sibling distance=30mm},
  level 2/.style={sibling distance=15mm}]
\node {$x_1$}
  child {node {$x_2$}
    child {node {$z_{00}$} edge from parent node[lab] {$0$}}
    child {node {$z_{01}$} edge from parent node[lab] {$1$}}
    edge from parent node[lab] {$0$}}
  child {node {$x_2$}
    child {node {$z_{10}$} edge from parent node[lab] {$0$}}
    child {node {$z_{11}$} edge from parent node[lab] {$1$}}
    edge from parent node[lab] {$1$}};
\node at (0,-2.4) {(b) a Littlestone tree of depth $3$};
\end{scope}
\end{tikzpicture}
\caption{Sequentialization. (a) The first epoch asks about $x_1$ and $x_2$ together; the second asks one question, which depends on both answers. (b) Asking the questions of each tuple one at a time, in a fixed order, gives a Littlestone tree with the same paths. It is a special Littlestone tree, in which the questions of one epoch do not depend on one another's answers: here $x_2$ is asked whatever the label of $x_1$. So the epoch dimension is at most the Littlestone dimension, and in general smaller.}
\label{fig:seqtree}
\end{figure}
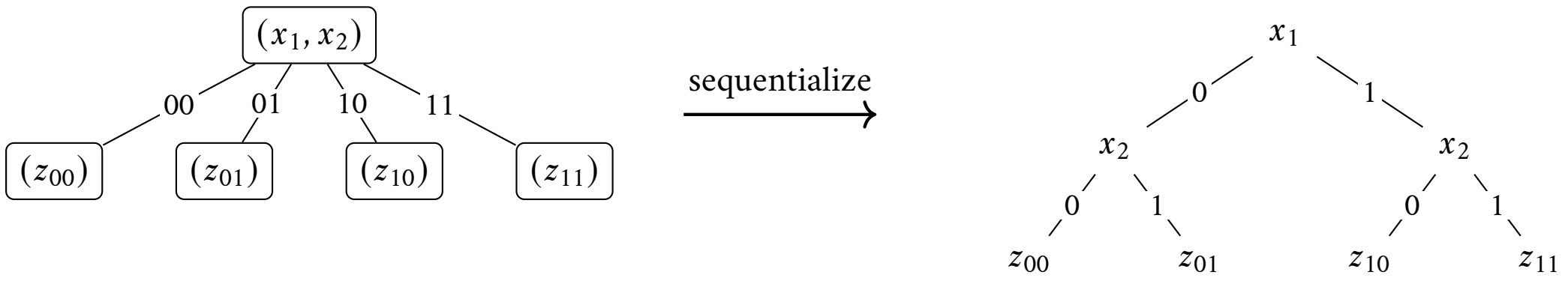

\textbf{Products.} For classes \(H_1\subseteq\{0,1\}^{Q_1}\) and
\(H_2\subseteq\{0,1\}^{Q_2}\) on disjoint sets, the \emph{product}
\(H_1\times H_2\) is the class on \(Q_1\sqcup Q_2\) of the functions
that agree with some \(h_1\in H_1\) on \(Q_1\) and with some
\(h_2\in H_2\) on \(Q_2\). VC and Littlestone dimension are additive
over products, and so is the epoch dimension.

\Prop{Additivity}

\label{prop:additive} For nonempty classes \(H_1\) and \(H_2\) on
disjoint sets and every \(n\ge0\),
\(\ED_n(H_1\times H_2)=\ED_n(H_1)+\ED_n(H_2)\).

\EndProp

\Pf{}

By induction on \(n\), for all pairs of nonempty classes at once; for
\(n=0\) both sides are \(0\). Let \(n\ge1\) and \(V=H_1\times H_2\). A
finite set \(A\) is shattered by \(V\) iff \(A_1=A\cap Q_1\) is
shattered by \(H_1\) and \(A_2=A\cap Q_2\) by \(H_2\), since the labels
on \(Q_1\) and on \(Q_2\) are chosen independently. A labeling \(\mu\)
of \(A\) is a pair \((\mu_1,\mu_2)\) of labelings of \(A_1\) and
\(A_2\), and
\(V_{A\to\mu}=(H_1)_{A_1\to\mu_1}\times(H_2)_{A_2\to\mu_2}\), whose
factors are nonempty when \(A\) is shattered. By the induction
hypothesis, \(\ED_{n-1}(V_{A\to\mu})\) is the sum of the dimensions of
the two factors, and its minimum over \(\mu\) is the sum of the minima
over \(\mu_1\) and over \(\mu_2\). So the term of Proposition
\ref{prop:recursion} for \(A\) and \(V\) is the sum of the terms for
\(A_1\) and \(H_1\) and for \(A_2\) and \(H_2\), and since \(A_1\) and
\(A_2\) range independently over the sets shattered by \(H_1\) and
\(H_2\), the supremum over \(A\) is the sum of the two suprema. \EndPf

\section{Examples}\label{sec:examples}

Table \ref{tab:examples} collects the dimensions of our examples, and
the rest of this section computes their epoch dimensions explicitly.

\begin{table}[ht]
\centering
\begin{tabular}{@{}lccc@{}}
\toprule
class $H$ & $\VC(H)$ & $\Ldim(H)$ & $\ED_n(H)$\\
\midrule
thresholds on $[0,1]$ & $1$ & $\infty$ & $n$\\
unions of $K$ intervals & $2K$ & $\infty$ & $2Kn$\\
cube $\{0,1\}^D$, $|D|=d$ & $d$ & $d$\\
half-spaces in $\mathbb{R}^d$ & $d+1$ & $\infty$ & $dn+1$\\
thresholds $\times$ cube & $d+1$ & $\infty$ & $n+d$\\
\bottomrule
\end{tabular}
\caption{The dimensions of our examples, for $n\ge1$. The last row is the product of the thresholds on $[0,1]$ and the cube.}
\label{tab:examples}
\end{table}

\Ex{Thresholds}

\label{ex:thresholds} Let \(Q=[0,1]\) and \(H=\{h_\theta\}\) with
\(h_\theta(x)=\mathbf{1}[x\ge\theta]\), so that \(\VC(H)=1\) and
\(\Ldim(H)=\infty\). Then \(\ED_n(H)=n\). The upper bound is Proposition
\ref{prop:bounds}. For the lower bound, let
\(V_{(a,b]}=\{h_\theta:a<\theta\le b\}\). The midpoint \(m\) of
\((a,b]\) is shattered by \(V_{(a,b]}\), leaving \(V_{(a,m]}\) or
\(V_{(m,b]}\), so by Proposition \ref{prop:recursion} and induction,
\(\ED_n(V_{(a,b]})\ge n\): binary search, one question per epoch, as in
Figure \ref{fig:shatter}(a).

\EndEx

\Ex{Unions of intervals}

\label{ex:intervals} For \(K\ge1\), let \(H\) be the unions of at most
\(K\) closed intervals on \(Q=[0,1]\), so that \(\VC(H)=2K\) and
\(\Ldim(H)=\infty\). Then \(\ED_n(H)=2Kn\); for intervals, \(2n\). The
upper bound is Proposition \ref{prop:bounds}. For the lower bound, which
runs \(2K\) binary searches side by side, take disjoint intervals
\(J_1<J_2<\dots<J_{2K}\) in \([0,1]\), and let \(V\) be the unions of
intervals \([a_i,b_i]\), \(i=1,\dots,K\), with each \(a_i\) ranging over
an interval \(I_{2i-1}\subseteq J_{2i-1}\) and each \(b_i\) over an
interval \(I_{2i}\subseteq J_{2i}\), all of positive length. On
\(J_{2i-1}\) every \(h\in V\) is the threshold at \(a_i\), on \(J_{2i}\)
the reversed threshold at \(b_i\), and the \(2K\) endpoints vary
independently. So the midpoints of \(I_1,\dots,I_{2K}\) are shattered by
\(V\), and each labeling of them confines every endpoint to one half of
its interval, leaving a class of the same form. By Proposition
\ref{prop:recursion} and induction, \(\ED_n(V)\ge2Kn\).

\EndEx

\Ex{Cubes}

\label{ex:cube} For the class \(\{0,1\}^D\) of all labelings of \(d\)
points, \(\ED_n=d\) for all \(n\): \(D\) is shattered, and by Lemma
\ref{lem:norepeat} no path carries more than \(d\) points.

\EndEx

\textbf{Products.} By Proposition \ref{prop:additive}, thresholds on two
disjoint copies of \([0,1]\), and thresholds on \([0,1]\) together with
all labelings of one further point, both have VC dimension \(2\) and
Littlestone dimension \(\infty\), but epoch dimensions \(2n\) and
\(n+1\). So for \(n\ge2\) the epoch dimension is not a function of the
VC and Littlestone dimensions.

\textbf{Half-spaces.} Let \(\mathcal{H}_d\) be the half-spaces
\(h_q(x)=\mathbf{1}[w\cdot x+b\ge0]\) on \(\mathbb{R}^d\), for
\(q=(w,b)\in\mathbb{R}^{d+1}\), and write
\(\mathcal{H}_U=\{h_q:q\in U\}\) for a set \(U\) of parameters;
\(\VC(\mathcal{H}_d)=d+1\) and \(\Ldim(\mathcal{H}_d)=\infty\).

\Lem{Pinned half-spaces}

\label{lem:radon} If every half-space in \(V\subseteq\mathcal{H}_d\)
gives a point \(x\) the same label \(b\), then \(\VC(V)\le d\).

\EndLem

\Pf{}

If \(V\) shattered \(y_1,\dots,y_{d+1}\) (all different from \(x\)),
Radon's theorem {[}Rad21{]} would split \(x,y_1,\dots,y_{d+1}\) into
parts \(P\ni x\) and \(P'\) with intersecting convex hulls. A witness
for the labeling giving the \(y_i\) in \(P\) the label \(b\) and those
in \(P'\) the label \(1-b\) would put these hulls in the disjoint convex
sets \(h^{-1}(b)\) and \(h^{-1}(1-b)\). \EndPf

\Lem{Open sets of half-spaces}

\label{lem:openhs} If \(U\subseteq\mathbb{R}^{d+1}\) is nonempty and
open, then \(\ED_m(\mathcal{H}_U)\ge dm\).

\EndLem

\Pf{}

Pick \(q=(w,b)\in U\) with \(w\ne0\) and \(d\) affinely independent
points \(x_1,\dots,x_d\) on the hyperplane \(w\cdot x+b=0\). The
functionals \(q'\mapsto(x_i,1)\cdot q'\) vanish at \(q\) and are
linearly independent, so near \(q\) they take every pattern of strict
signs: each labeling \(\mu\) of \(\{x_1,\dots,x_d\}\) is realized by all
of \(\mathcal{H}_{U_\mu}\) for a nonempty open \(U_\mu\subseteq U\).
Induct with Proposition \ref{prop:recursion}. \EndPf

\Prop{Half-spaces}

\label{prop:halfspaces} \(\ED_n(\mathcal{H}_d)=dn+1\) for every
\(n\ge1\).

\EndProp

\Pf{}

Upper bound: in Proposition \ref{prop:recursion}, a nonempty shattered
\(A\) has at most \(d+1\) points, and each \((\mathcal{H}_d)_{A\to\mu}\)
has VC dimension at most \(d\) by Lemma \ref{lem:radon}, hence
\(\ED_{n-1}\le d(n-1)\) by Proposition \ref{prop:bounds}; the empty set
gives \(\ED_{n-1}(\mathcal{H}_d)\), which is smaller by induction. Lower
bound: \(d+1\) affinely independent points are shattered, each labeling
by an open set of parameters, and Lemma \ref{lem:openhs} applies below.
\EndPf

The first epoch gets the VC dimension, and every later epoch only the
\(d\) parameters of the boundary (Figure \ref{fig:halfplanes}). For
half-planes, \(2n+1\) lies strictly between the bounds \(\max(3,n)\) and
\(3n\) of Proposition \ref{prop:bounds} for \(n\ge2\). The gain can also
be larger. By Proposition \ref{prop:additive}, the product \(H\) of the
thresholds on \([0,1]\) and the cube \(\{0,1\}^D\) with \(|D|=d\) has
\(\ED_n(H)=n+d\).

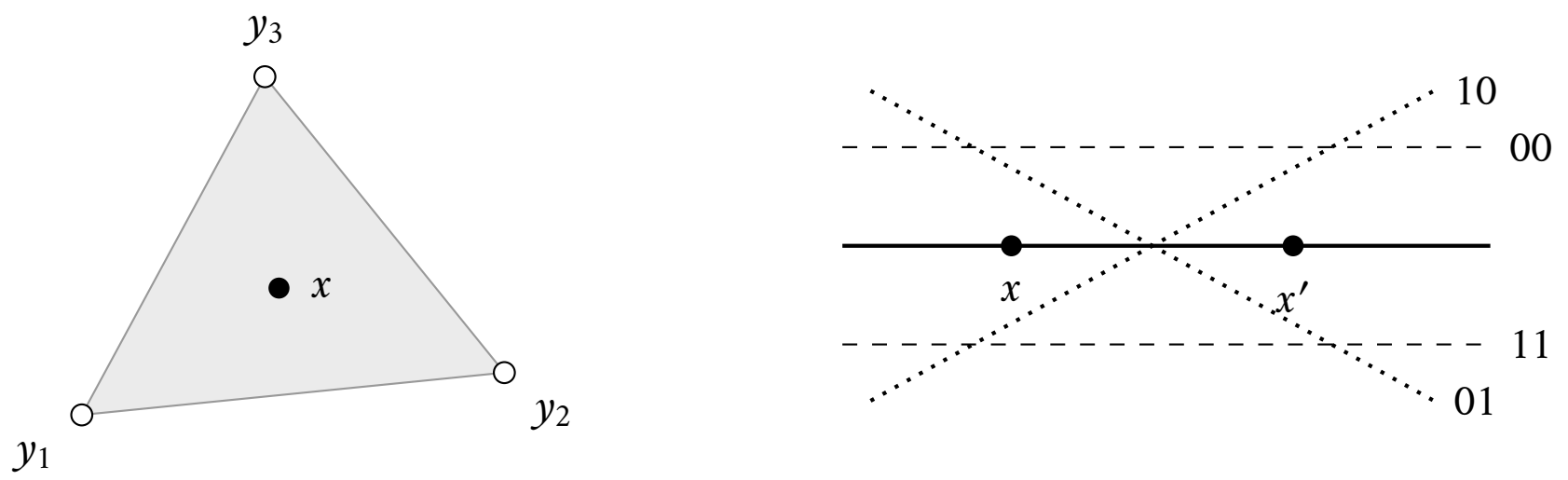
\begin{figure}[ht]
\centering
\begin{tikzpicture}[every node/.style={font=\footnotesize},
  pt/.style={circle,fill,inner sep=1.5pt},
  opt/.style={circle,draw,fill=white,inner sep=1.5pt}]
\begin{scope}
\coordinate (y1) at (0,0); \coordinate (y2) at (3,0.3); \coordinate (y3) at (1.3,2.4);
\fill[black!8] (y1) -- (y2) -- (y3) -- cycle;
\draw[black!40] (y1) -- (y2) -- (y3) -- cycle;
\node[opt,label=below left:{$y_1$}] at (y1) {};
\node[opt,label=below right:{$y_2$}] at (y2) {};
\node[opt,label=above:{$y_3$}] at (y3) {};
\node[pt,label=right:{$x$}] at (1.4,0.9) {};
\node[align=center] at (1.5,-1.0) {(a) a known label rules out\\ a labeling of every three points};
\end{scope}
\begin{scope}[xshift=6.6cm,yshift=1.2cm]
\draw[thick] (-1.2,0) -- (3.4,0);
\draw[dashed] (-1.2,0.7) -- (3.4,0.7) node[right] {$00$};
\draw[dashed] (-1.2,-0.7) -- (3.4,-0.7) node[right] {$11$};
\draw[dotted,thick] (-1,-1.1) -- (3,1.1) node[right] {$10$};
\draw[dotted,thick] (-1,1.1) -- (3,-1.1) node[right] {$01$};
\node[pt,label={[label distance=1pt]below:{$x$}}] at (0,0) {};
\node[pt,label={[label distance=1pt]below:{$x'$}}] at (2,0) {};
\node[align=center] at (1.1,-2.2) {(b) two points on a boundary line\\ take all four labelings nearby};
\end{scope}
\end{tikzpicture}
\caption{Why half-planes get three points in the first epoch and two in every later one. (a) Once the label of a point $x$ is known, say $1$, some labeling of any three further points is ruled out. Here $x$ lies in the triangle $y_1y_2y_3$, and no half-plane gives $x$ the label $1$ and $y_1,y_2,y_3$ the label $0$, because the set labeled $0$ is convex; Radon's theorem gives such an obstruction for every three points (Lemma \ref{lem:radon}). (b) Two points $x,x'$ on the boundary line of a half-plane, labeled $1$ above each line, take all four labelings from nearby half-planes: shift the line up ($00$) or down ($11$), or rotate it about the midpoint of $xx'$ ($10$ and $01$). Each labeling leaves an open set of half-planes, so the next epoch can do the same (Lemma \ref{lem:openhs}).}
\label{fig:halfplanes}
\end{figure}

\section{A lower bound for learning under one law}\label{sec:lower}

\Defn{The history game}

\label{def:game} Fix \(H\) with \(1\le\VC(H)=d<\infty\), a bound
\(n\ge1\), and a horizon \(T\ge2\) known to the learner. An adversary
chooses a law \(c\in H\), possibly at random, and then divides rounds
\(1,\dots,T\) into at most \(n\) consecutive epochs, starting at rounds
\(t_1=1<t_2<\cdots\), choosing at the start of each its length and a
distribution \(D_j\) on \(Q\), after seeing everything so far. At each
round of epoch \(j\), a point \(x_t\sim D_j\) is drawn independently of
the past; the learner predicts its label, possibly at random, and then
sees \(c(x_t)\). The learner is not necessarily told where the epochs
begin; in the variant with \emph{announced boundaries}, it is. We write
\(k_j\) for the length of epoch \(j\), after the last epoch is cut off
at round \(T\); \(M_j\) for the learner's mistakes in it, with \(M_j=0\)
if there is no \(j\)-th epoch; and \(\mathcal{F}_j\) for the law \(c\),
everything that happened before round \(t_j\), and the adversary's
choices for epoch \(j\). Finally, \(\Opt(H;n,T)\) is the infimum over
learners of the supremum over adversaries of \(\E\big[\sum_jM_j\big]\),
and \(\Opt^\circ(H;n,T)\le\Opt(H;n,T)\) is the same quantity with
announced boundaries. We assume, as usual, the measurability needed for
these expectations to be defined.

\EndDefn

\Thm{Aimed views}

\label{thm:lower} Let \(1\le\VC(H)<\infty\), \(n\ge1\), and
\(N=\ED_n(H)\). Some random adversary in the history game, the same for
every learner, forces every learner, even one told where the epochs
begin, to make
\[\E\Big[\sum_jM_j\Big]\;\ge\;\begin{cases}\frac{1-e^{-m}}{2}\,N&\text{if }T\ge mN\text{ for an integer }m\ge1,\\[3pt]\frac{1-1/e}{2}\,T&\text{if }T<N.\end{cases}\]
In particular, \(\E\big[\sum_jM_j\big]\ge\frac{1-1/e}{2}\min(T,N)\) for
every \(T\).

\EndThm

\Pf{}

\(N\) is finite by Proposition \ref{prop:bounds}. If \(T<N\), replace
\(N\) by \(T\) and let \(m=1\); by Proposition \ref{prop:recursion},
\(H\) still shatters an epoch tree of depth \(n\) with \(N\) points.
Take such a tree. The adversary walks down a random path: at the node
reached at level \(j\), with tuple \(A_j\), it picks a labeling
\(\lambda_j\) of \(A_j\) uniformly at random, independently of the
earlier choices, and it lets \(c\in H\) realize the path. Epoch \(j\)
has length \(m|A_j|\) and the uniform distribution on \(A_j\); epochs
with empty tuples are skipped, and the last runs on to round \(T\),
which fits since \(mN\le T\). (Whether a node's epoch is the last
nonempty one on the path does not depend on the node's labeling, since
every path carries exactly \(N\) points; so the length of each epoch is
fixed before its labels are drawn.)

Before epoch \(j\) the learner has seen only points of
\(A_1,\dots,A_{j-1}\), which are disjoint from \(A_j\) by Lemma
\ref{lem:norepeat}, with the labels \(\lambda_1,\dots,\lambda_{j-1}\).
At a round where a point \(x\in A_j\) appears for the first time in the
epoch, its label \(\lambda_j(x)\) is therefore independent of everything
the learner has seen, including the epoch's earlier examples, which show
other coordinates of \(\lambda_j\); so the learner errs with probability
exactly \(\frac12\). The epoch's \(k_j\ge ma\) uniform draws, where
\(a=|A_j|\), show \(a\big(1-(1-1/a)^{k_j}\big)\ge(1-e^{-m})\,a\)
distinct points in expectation. So
\(\E[M_j]\ge\frac{1-e^{-m}}{2}\E|A_j|\), and summing over the path,
which carries \(N\) points, gives the bound. Nothing in the adversary
depends on the learner. \EndPf

\Rem{Points presented directly}

\label{rem:direct} Suppose that at the start of each epoch the adversary
may fix the sequence of its points instead of a distribution. Against a
deterministic learner, the walk above then forces \(\ED_n(H)\) mistakes
when \(T\ge\ED_n(H)\): the adversary presents the points of \(A_j\) in
turn and gives each the label opposite to the learner's prediction,
which some path allows, since \(A_j\) is shattered by the hypotheses
realizing the paths through its node; as the learner is deterministic,
the adversary can compute the resulting path, and so the law, in
advance. Against a randomized learner, uniformly random labels force
\(\frac12\ED_n(H)\) expected mistakes, which the bound of Theorem
\ref{thm:lower} approaches as \(T/\ED_n(H)\) grows. So the factor
\(1-1/e\) there is the price not of sampling but of a horizon as short
as \(\ED_n(H)\).

\EndRem

\section{A matching upper bound}\label{sec:upper}

We now give a learner that attains Theorem \ref{thm:lower} up to a
logarithmic factor, first when it is told where the epochs begin and
then when it is not. It combines three classical ingredients: the
one-inclusion graph {[}HLW94{]}, Littlestone's standard optimal
algorithm {[}Lit88{]}, and fixed-share aggregation {[}HW98{]}.

\subsection{The layered SOA}\label{sec:lsoa}

\Defn{One-inclusion graph predictor [HLW94]}

\label{def:oig} For a finite set \(U\) and a class
\(C\subseteq\{0,1\}^U\), the \emph{one-inclusion graph} \(G(C)\) has the
labelings in \(C\) as vertices, and an edge in \emph{direction} \(x\)
between two labelings that differ exactly at the point \(x\). An
orientation of its edges is \emph{admissible} if every vertex has at
most \(\VC(C)\) outgoing edges. Given an admissible orientation of
\(G(C)\) and the labels of all points of \(U\) but one, \(x\), the
\emph{candidates} are the labelings in \(C\) that agree with the known
labels; there are at most two, and they differ at \(x\). The
\emph{one-inclusion graph predictor} predicts the label at \(x\) of the
candidate if only one remains, and otherwise the label at \(x\) of the
endpoint to which the edge between the two candidates points.

\EndDefn

Haussler, Littlestone, and Warmuth showed that every subgraph of
\(G(C)\) has at most \(\VC(C)\) times as many edges as vertices, so that
admissible orientations exist {[}HLW94{]}. Because the orientation
depends on \(U\) alone, the predictor's errors can be counted by
symmetry.

\Lem{Leave one out}

\label{lem:loo} Let \(x_1,\dots,x_{m+1}\) be random points whose joint
conditional distribution given a \(\sigma\)-field \(\mathcal{G}\) is
exchangeable, and let \(B\) be a random finite set determined by
\(\mathcal{G}\) and the multiset \(\{x_1,\dots,x_{m+1}\}\). Then the
conditional probability given \(\mathcal{G}\) that \(x_{m+1}\in B\) and
\(x_{m+1}\ne x_s\) for all \(s\le m\) is at most
\(\E\big[|B|\;\big|\;\mathcal{G}\big]/(m+1)\).

\EndLem

\Pf{}

Let \(E_s\) be the event that \(x_s\in B\) and \(x_s\) occurs only once
among \(x_1,\dots,x_{m+1}\). Exchanging \(x_s\) and \(x_{m+1}\) changes
neither \(B\) nor the conditional distribution, so all the \(E_s\) have
the same conditional probability; and \(\sum_s\mathbf{1}[E_s]\le|B|\).
\EndPf

Littlestone's \emph{standard optimal algorithm} (SOA) keeps the version
space \(V\) and on a query \(x\) predicts a label \(b\) that maximizes
\(\Ldim(V_{x\to b})\), with \(\Ldim(\emptyset)=-1\); every mistake
lowers \(\Ldim(V)\), so it makes at most \(\Ldim(H)\) mistakes
{[}Lit88{]}. The \emph{layered} SOA instead predicts the label under
which the epoch dimension for the remaining epochs stays larger. When
the two labels leave the same value, it falls back on the one-inclusion
graph of the labelings that keep the future fully open; in the last
epoch every value is \(0\), and it is the one-inclusion graph predictor.
The key lemma says that labelings of an epoch's points that leave the
future rich must be simple now.

\Lem{High-value labelings}

\label{lem:highvalue} Let \(V\ne\emptyset\), \(r\ge0\), and
\(U\subseteq Q\) finite. For a labeling \(\lambda\) of \(U\) realized by
\(V\), let \(f(\lambda)=\ED_r(V_{U\to\lambda})\), the dimension that
remains for \(r\) further epochs if the points of \(U\) are labeled by
\(\lambda\). Then for every integer \(\tau\) for which the set on the
left is nonempty,
\[\VC\big(\{\lambda\in V|_U:f(\lambda)\ge\tau\}\big)\;\le\;\ED_{r+1}(V)-\tau.\]

\EndLem

\Pf{}

If \(A\subseteq U\) is shattered by the set on the left, every labeling
\(\mu\) of \(A\) extends to some \(\lambda\in V|_U\) with
\(f(\lambda)\ge\tau\), and
\(V_{A\to\mu}\supseteq V_{U\to\lambda}\ne\emptyset\), so
\(\ED_r(V_{A\to\mu})\ge\tau\) by monotonicity (Proposition
\ref{prop:bounds}). So \(V\) shatters \(A\), and Proposition
\ref{prop:recursion} gives \(\ED_{r+1}(V)\ge|A|+\tau\). \EndPf

\Defn{Layered SOA}

\label{def:lsoa} Fix, for every finite \(U\subseteq Q\) and every
\(C\subseteq\{0,1\}^U\), an admissible orientation of \(G(C)\)
(Definition \ref{def:oig}). The \emph{layered SOA} (LSOA) knows a bound
\(n\) on the number of epochs and is told the boundaries as they occur.
At the start of epoch \(j\), let \(V\) be the version space and
\(r=n-j\), and define \(f\) as in Lemma \ref{lem:highvalue}. At a round
of epoch \(j\) with query \(x\), if \(x\) has occurred before, LSOA
predicts its known label. Otherwise let \(U'\) be the set of the epoch's
earlier points, \(U=U'\cup\{x\}\), and \(\lambda_b\) the labeling of
\(U\) that extends the known labels of \(U'\) by \(x\mapsto b\). LSOA
predicts as follows (Figure \ref{fig:lsoa}).

\begin{itemize}
\tightlist
\item
  \emph{Determined.} If only one \(\lambda_b\) lies in \(V|_U\), predict
  that \(b\).
\item
  \emph{Decisive.} Otherwise, if \(f(\lambda_0)\ne f(\lambda_1)\),
  predict the label with the larger value.
\item
  \emph{Tie.} Otherwise, with \(\tau=f(\lambda_0)=f(\lambda_1)\),
  predict the label of the endpoint to which the edge between
  \(\lambda_0\) and \(\lambda_1\) points in \(G(C)\), where
  \(C=\{\lambda\in V|_U:f(\lambda)\ge\tau\}\).
\end{itemize}

\EndDefn

\begin{figure}[ht]
\centering
\begin{tikzpicture}[every node/.style={font=\footnotesize},
  v/.style={draw,rounded corners=2pt,minimum width=7mm,minimum height=5mm,inner sep=1pt},
  g/.style={v,draw=black!30,text=black!40}]
\begin{scope}
\node[g] (a00) at (0,0) {$00$};
\node[g] (a01) at (2.6,0) {$01$};
\node[v,thick] (a10) at (0,1.6) {$10$};
\node[v,thick] (a11) at (2.6,1.6) {$11$};
\draw[black!30] (a00) -- (a01) (a00) -- (a10) (a01) -- (a11);
\draw (a10) -- node[below,font=\scriptsize] {$x$} (a11);
\node[above=1pt] at (a10.north) {$f=2$};
\node[above=1pt] at (a11.north) {$f=1$};
\node at (1.3,-0.75) {(a) decisive: predict $0$};
\end{scope}
\begin{scope}[xshift=6.2cm]
\node[g] (b00) at (0,0) {$00$};
\node[g] (b01) at (2.6,0) {$01$};
\node[v,thick] (b10) at (0,1.6) {$10$};
\node[v,thick] (b11) at (2.6,1.6) {$11$};
\draw[black!30] (b00) -- (b01) (b00) -- (b10) (b01) -- (b11);
\draw[->,thick] (b10) -- node[below,font=\scriptsize] {$x$} (b11);
\node[above=1pt] at (b10.north) {$f=1$};
\node[above=1pt] at (b11.north) {$f=1$};
\node at (1.3,-0.75) {(b) tie: predict $1$};
\end{scope}
\end{tikzpicture}
\caption{The layered SOA at a query $x$, after the epoch has shown that a point $u$ has label $1$. Vertices are labelings of $U=\{u,x\}$, written as the labels of $u$ and $x$; those inconsistent with what is known are gray. The two candidates differ at $x$, and $f$ is the epoch dimension that remains for the later epochs if a candidate is the truth. (a) If the values differ, predict the label whose candidate leaves more; a mistake then lowers what remains. (b) If they are equal, follow an admissible orientation of the one-inclusion graph of the labelings whose value is at least as large, and predict the label of the vertex the edge points to.}
\label{fig:lsoa}
\end{figure}
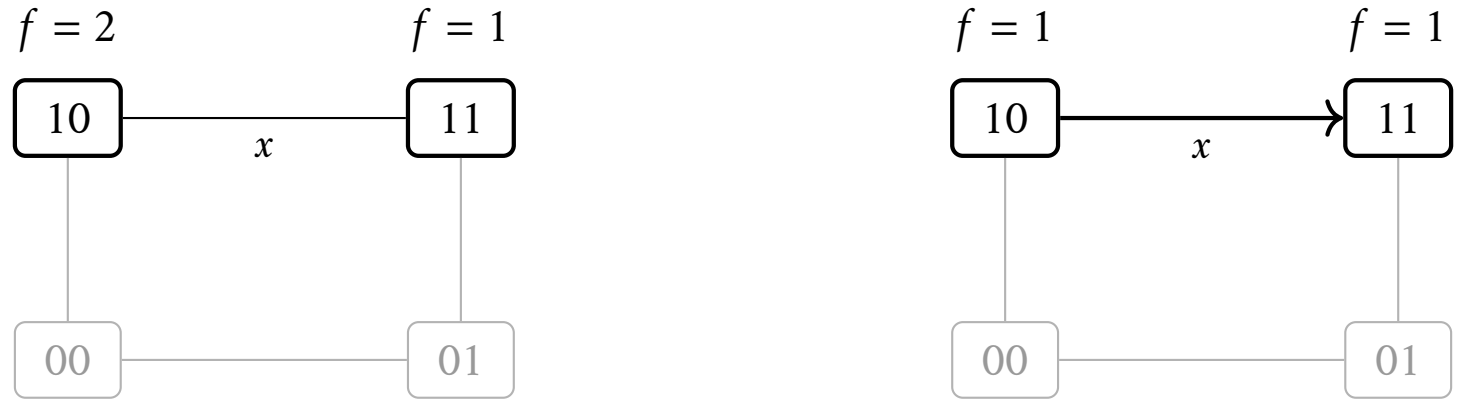

The analysis measures each epoch by how much of the epoch dimension it
uses up (Figure \ref{fig:pe1}).

\Lem{One epoch of LSOA}

\label{lem:lsoa-epoch} Suppose that the boundaries are announced. Let
\(V_{t_j}\) be the version space at the start of epoch \(j\), or at the
end of the game if there is no \(j\)-th epoch, let
\(\Phi_j=\ED_{n-j+1}(V_{t_j})\), and let
\(\Delta_j=\Phi_j-\Phi_{j+1}\ge0\). Then LSOA satisfies, for every epoch
\(j\) that occurs,
\[\E[M_j\mid\mathcal{F}_j]\;\le\;(2+\ln k_j)\,\E[\Delta_j\mid\mathcal{F}_j].\]

\EndLem

\Pf{}

Condition on \(\mathcal{F}_j\), which fixes \(c\), \(k_j\), \(D_j\), and
\(V=V_{t_j}\), and let \(r=n-j\). (Here we use that the adversary fixes
\(k_j\) when the epoch begins; were the epoch to end at a stopping time
of its own sample, its points would no longer be exchangeable given
\(\mathcal{F}_j\).) For a set \(U\) of the epoch's points, let
\(g(U)=f(c|_U)\), the value that the truth leaves for the later epochs.
It is non-increasing in \(U\), \(g(\emptyset)=\ED_r(V)\le\Phi_j\), and
\(g(S)=\Phi_{j+1}\) for the set \(S\) of all the epoch's points.
Determined rounds cost nothing. In a decisive round both values are at
most \(g(U')\), and a mistake means that the truth has the smaller one,
so \(g\) drops by at least \(1\): the epoch has at most
\(g(\emptyset)-g(S)\le\Delta_j\) decisive mistakes. In a tie round
\(\tau=g(U)\), so a mistake at the epoch's \((m+1)\)-st round requires
its point to occur only once among the epoch's first \(m+1\) points,
which form \(U\), and to be the direction of an edge pointing away from
\(c|_U\) in \(G(C_U)\), where
\(C_U=\{\lambda\in V|_U:f(\lambda)\ge g(U)\}\). Given \(\mathcal{F}_j\),
the set \(B\) of these directions is determined by \(U\), and by Lemma
\ref{lem:highvalue}
\[|B|\;\le\;\VC(C_U)\;\le\;\ED_{r+1}(V)-g(U)\;\le\;\Phi_j-g(S)\;=\;\Delta_j.\]
The epoch's points are i.i.d. given \(\mathcal{F}_j\), so by Lemma
\ref{lem:loo} a tie mistake occurs at this round with conditional
probability at most \(\E[\Delta_j\mid\mathcal{F}_j]/(m+1)\). Summing
over \(m<k_j\) and adding the decisive mistakes gives the bound, since
\(1+\sum_{m=1}^{k}\frac1m\le2+\ln k\). \EndPf

\Thm{The layered SOA}

\label{thm:lsoa} Let \(\VC(H)=d\ge1\). Against every adversary in the
history game that announces the epoch boundaries, LSOA satisfies
\[\E\Big[\sum_jM_j\Big]\;\le\;(2+\ln T)\ED_n(H).\]

\EndThm

\Pf{}

Whether epoch \(j\) occurs, and its length \(k_j\le T\), are determined
by \(\mathcal{F}_j\), so Lemma \ref{lem:lsoa-epoch} gives
\(\E[M_j]\le(2+\ln T)\E[\Delta_j]\). Summing over \(j\), the
\(\Delta_j\) telescope to at most \(\Phi_1=\ED_n(H)\). \EndPf

\subsection{Hidden boundaries}\label{sec:hidden}

LSOA must be told where each epoch begins, because the values it
compares depend on how many epochs remain. The remedy is to run a copy
of LSOA for each round at which the current epoch might have begun and
each number of epochs that might remain, and to aggregate the copies by
fixed share (Figure \ref{fig:hidden}).

\Defn{Fixed share [HW98]}

\label{def:fixedshare} Suppose experts \(i=1,\dots,K\) predict \(1\) at
round \(t\) with probabilities \(p_t(i)\). Given weights \(w_t\), the
aggregate predicts \(1\) with probability \(\sum_iw_t(i)p_t(i)\); expert
\(i\) suffers the loss \(\ell_t(i)=|p_t(i)-y_t|\), its probability of
error, and the aggregate suffers \(\hat\ell_t=\sum_iw_t(i)\ell_t(i)\).
The \emph{fixed-share} rule, with a prior \(w_1\) whose weights are all
positive, a learning rate \(\eta>0\), and a share rate
\(\gamma\in[0,1)\), updates
\[v_{t+1}(i)=\frac{w_t(i)e^{-\eta\ell_t(i)}}{\sum_{i'}w_t(i')e^{-\eta\ell_t(i')}},\qquad w_{t+1}(i)=(1-\gamma)\,v_{t+1}(i)+\gamma\,w_1(i).\]
With \(\gamma=0\) it is \emph{exponential weights}.

\EndDefn

The share step keeps every weight at least \(\gamma w_1(i)\), so the
aggregate can quickly follow an expert that starts to do well.

\Lem{Interval bound for fixed share [HW98]}

\label{lem:fs} For all losses \(\ell_t(i)\in[0,1]\), every interval
\([s,s']\subseteq[1,T]\), and every expert \(i\),
\[\sum_{t=s}^{s'}\hat\ell_t\;\le\;\frac{\eta\sum_{t=s}^{s'}\ell_t(i)+\ln\frac{1}{\omega_s(i)}+(s'-s+1)\ln\frac{1}{1-\gamma}}{1-e^{-\eta}},\]
where \(\omega_1(i)=w_1(i)\) and \(\omega_s(i)=\gamma\,w_1(i)\) for
\(s\ge2\).

\EndLem

\Pf{}

Since \(e^{-\eta\ell}\le1-(1-e^{-\eta})\ell\) for \(\ell\in[0,1]\), the
denominator of \(v_{t+1}(i)\) is at most
\(\exp\big(-(1-e^{-\eta})\hat\ell_t\big)\), so
\(\ln w_{t+1}(i)\ge\ln w_t(i)+\ln(1-\gamma)-\eta\ell_t(i)+(1-e^{-\eta})\hat\ell_t\).
Sum over \(t=s,\dots,s'\), and use \(w_{s'+1}(i)\le1\) and
\(w_s(i)\ge\omega_s(i)\). \EndPf

\Thm{Hidden boundaries}

\label{thm:lsoa-hidden} Let \(1\le\VC(H)<\infty\). For
\(1\le\tau,r\le T\), let \(E_{\tau,r}\) predict \(0\) before round
\(\tau\) and, from round \(\tau\) on, follow LSOA as if an epoch began
at \(\tau\), with \(r-1\) epochs to follow, starting from the version
space of rounds \(1,\dots,\tau-1\). Aggregate the \(T^2\) experts
\(E_{\tau,r}\) by fixed share (Definition \ref{def:fixedshare}) with the
uniform prior, \(\eta=1\), and \(\gamma=1/T\), and aggregate the result
with SOA, if \(\Ldim(H)<\infty\), by exponential weights with \(\eta=1\)
and the uniform prior. This learner does not depend on \(n\), and for
every \(n\ge1\), against every adversary that uses at most \(n\) epochs
and does not announce their boundaries,
\[\E\Big[\sum_jM_j\Big]\;\le\;12\,(1+\ln T)\ED_n(H).\]

\EndThm

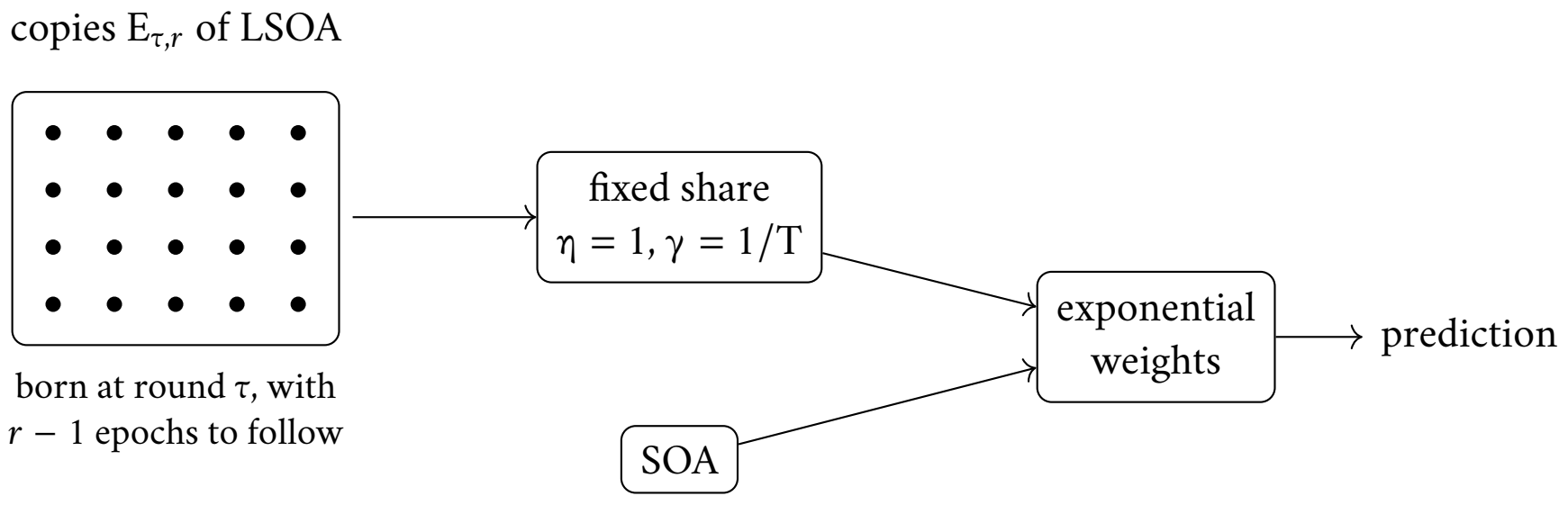
\begin{figure}[ht]
\centering
\begin{tikzpicture}[every node/.style={font=\footnotesize},
  box/.style={draw,rounded corners=3pt,align=center,inner sep=4pt}]
\foreach \i in {0,...,4} {\foreach \j in {0,...,3} {\fill (0.45*\i,-0.42*\j) circle (1.6pt);}}
\draw[rounded corners=3pt] (-0.3,0.3) rectangle (2.1,-1.56);
\node[align=center] at (0.9,0.75) {copies $E_{\tau,r}$ of LSOA};
\node[align=center,font=\scriptsize] at (0.9,-2.05) {born at round $\tau$, with\\ $r-1$ epochs to follow};
\node[box] (fs) at (4.6,-0.62) {fixed share\\ $\eta=1$, $\gamma=1/T$};
\node[box] (soa) at (4.6,-2.4) {SOA};
\node[box] (ew) at (8.1,-1.5) {exponential\\ weights};
\node (out) at (10.4,-1.5) {prediction};
\draw[->] (2.2,-0.62) -- (fs);
\draw[->] (fs) -- (ew);
\draw[->] (soa) -- (ew);
\draw[->] (ew) -- (out);
\end{tikzpicture}
\caption{The learner of Theorem \ref{thm:lsoa-hidden}. During each epoch, the copy of LSOA born at the epoch's first round with the right number of epochs to follow behaves exactly as LSOA would if told the boundaries; fixed share finds it after $O(\log T)$ mistakes. A second layer hedges against $n$ exceeding the Littlestone dimension, where SOA alone is optimal.}
\label{fig:hidden}
\end{figure}

\Pf{}

At most \(T\) epochs fit into \(T\) rounds, so the adversary uses at
most \(n'=\min(n,T)\) epochs, and \(\ED_{n'}(H)\le\ED_n(H)\); so we may
assume \(n\le T\). Then during epoch \(j\) the expert \(E_{t_j,n-j+1}\)
behaves exactly as LSOA does when told the boundaries, so by Lemma
\ref{lem:lsoa-epoch} it makes at most
\((2+\ln T)\E[\Delta_j\mid\mathcal{F}_j]\) expected mistakes in epoch
\(j\) given \(\mathcal{F}_j\). Write \(\alpha=e/(e-1)\) and
\(\Lambda=\ED_n(H)\). Apply Lemma \ref{lem:fs}, with \(K=T^2\) experts,
on each epoch to that expert:
\(\ln\frac1{\omega_s}\le\ln K+\ln T=3\ln T\) per epoch, and the share
terms add up to at most
\(T\ln\frac{1}{1-\gamma}\le\frac{T\gamma}{1-\gamma}\le2\) in all. Since
\(\sum_j\Delta_j\le\Phi_1=\Lambda\), the fixed-share layer makes at most
\(\alpha\big((2+\ln T)\Lambda+3n\ln T+2\big)\) expected mistakes. By
Lemma \ref{lem:fs} with \(\gamma=0\) and \(s=1\), and Littlestone's
bound for SOA, the top layer makes at most \(\alpha\) times the smaller
of this and \(\Ldim(H)\), plus \(\alpha\ln2\). If \(n\le\Ldim(H)\), then
\(\Lambda\ge n\) by Proposition \ref{prop:bounds}, and the fixed-share
bound is at most \(\alpha(4+4\ln T)\Lambda\); if \(n>\Ldim(H)\), then
\(\Lambda=\Ldim(H)\). Either way the total is at most
\(4\alpha^2(1+\ln T)\Lambda+\alpha\ln2\le12(1+\ln T)\Lambda\), since
\(\Lambda\ge1\). Nothing in the learner depends on \(n\). \EndPf

\subsection{Optimal learning under one law}\label{sec:main}

\Thm{One law}

\label{thm:main} Let \(1\le\VC(H)<\infty\) and \(n\ge1\). Then
\[\frac{1-1/e}{2}\,\min\big(T,\ED_n(H)\big)\;\le\;\Opt^\circ(H;n,T)\;\le\;\Opt(H;n,T)\;\le\;\min\big(\Ldim(H),\ 12\,(1+\ln T)\,\ED_n(H)\big),\]
and \(\Opt^\circ(H;n,T)\le(2+\ln T)\ED_n(H)\), attained by the layered
SOA.

\EndThm

\Pf{}

Theorems \ref{thm:lower}, \ref{thm:lsoa}, and \ref{thm:lsoa-hidden}, and
Littlestone's bound for SOA. \EndPf

So under one law, the best worst-case number of mistakes is the epoch
dimension, up to a factor \(O(\log T)\): the VC dimension for a single
epoch, and the Littlestone dimension for unboundedly many. The logarithm
can matter only when \(\Ldim(H)\) is much larger than \(\ED_n(H)\), but
then it cannot be removed in general. For thresholds, an adversary that
keeps aiming at the remaining uncertainty extracts in every epoch what a
fresh threshold would have cost (Figure \ref{fig:zoom}).

\Prop{Zooming}

\label{prop:zoom} Let \(H\) be the thresholds on \([0,1]\), so that
\(\ED_n(H)=n\). For every pattern of epoch lengths, a random law and a
choice of distributions that depends only on the labels seen so far
force every learner to make \(\E[M_j]\ge\frac12\ln\frac{k_j+2}{2}\) in
every epoch \(j\).

\EndProp

\Pf{}

Let \(\theta\) be uniform on \([0,1]\), and let \(D_j\) be uniform on
the gap \(G_j\) between the points seen before epoch \(j\) that contains
\(\theta\), which the labels seen so far determine. Given what the
learner has seen, \(\theta\) is uniform on \(G_j\), and all earlier
points lie outside \(G_j\), so, rescaled, each epoch is a fresh copy of
the problem. In it, after \(m\) examples, the next point falls into the
same one of the \(m+1\) gaps as \(\theta\) with probability \(2/(m+2)\);
given that, the two are independent and uniform on that gap, and the
best prediction errs with probability \(\E\min(u,1-u)=\frac14\) for
\(u\) uniform on \([0,1]\). Summing \(\frac{1}{2(m+2)}\) over \(m<k_j\)
gives the bound. \EndPf

\begin{figure}[ht]
\centering
\begin{tikzpicture}[x=11cm,y=1cm,every node/.style={font=\footnotesize},
  one/.style={circle,fill,inner sep=1.4pt},
  zero/.style={circle,draw,fill=white,inner sep=1.4pt}]
\def\th{0.62}
\foreach \y/\a/\b/\lab in {0/0/1/{epoch 1},-1.1/0.47/0.70/{epoch 2},-2.2/0.58/0.66/{epoch 3}} {
  \draw[black!35] (0,\y) -- (1,\y);
  \fill[black!12] (\a,\y-0.22) rectangle (\b,\y+0.22);
  \draw (\a,\y-0.22) -- (\a,\y+0.22) (\b,\y-0.22) -- (\b,\y+0.22);
  \node[anchor=east] at (-0.02,\y) {\lab};
}
\draw[dotted,thick] (\th,0.55) -- (\th,-2.6) node[below] {$\theta$};
\node[font=\scriptsize] at (0,0.42) {$0$};
\node[font=\scriptsize] at (1,0.42) {$1$};
\foreach \x in {0.08,0.31,0.47} {\node[zero] at (\x,0) {};}
\foreach \x in {0.70,0.88} {\node[one] at (\x,0) {};}
\foreach \x in {0.51,0.58} {\node[zero] at (\x,-1.1) {};}
\foreach \x in {0.66} {\node[one] at (\x,-1.1) {};}
\node[zero] at (0.60,-2.2) {};
\node[one] at (0.64,-2.2) {};
\node[anchor=west,font=\scriptsize] at (1.02,0) {$G_1=[0,1]$};
\node[anchor=west,font=\scriptsize] at (1.02,-1.1) {$G_2$};
\node[anchor=west,font=\scriptsize] at (1.02,-2.2) {$G_3$};
\end{tikzpicture}
\caption{The zooming adversary for thresholds (Proposition \ref{prop:zoom}). The threshold $\theta$ (dotted) is uniformly random. Each epoch's distribution is uniform on the gap $G_j$ between earlier points that contains $\theta$ (shaded); points to the right of $\theta$ are labeled $1$ (filled) and the others $0$ (open). Rescaled, every epoch is a fresh copy of the problem, so each costs every learner about $\frac12\ln k_j$ mistakes, although the law never changes.}
\label{fig:zoom}
\end{figure}
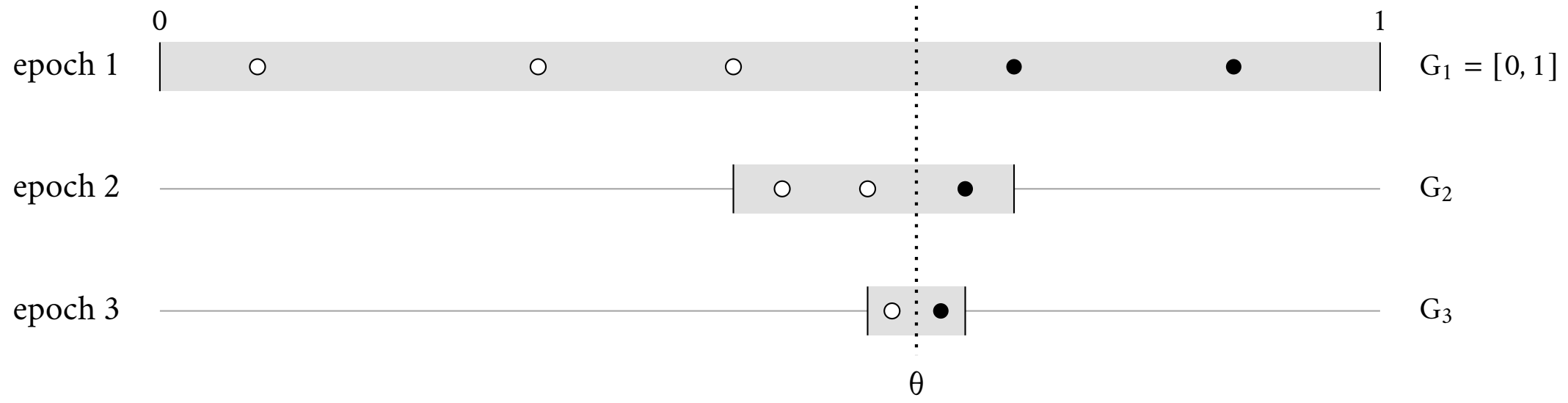

With \(n\) epochs of length \(T/n\), every learner thus makes
\(\Omega(n\log(T/n))\) expected mistakes on a class whose epoch
dimension is \(n\).

\section{Related work}\label{sec:related}

\textbf{Trees between shattered sets and Littlestone trees.} Closest to
epoch trees are the VCL trees of Bousquet, Hanneke, Moran, van Handel,
and Yehudayoff {[}BHM+21{]}. In a VCL tree the node at depth \(k\)
carries a tuple of \(k+1\) points and has a child for each labeling of
it, and the tree is shattered when every path is realized by a single
hypothesis. So a VCL tree of depth \(n\) is an epoch tree of depth \(n\)
whose widths grow as \(1,2,\dots,n\), shattered under one law. Bousquet,
Hanneke, Moran, Shafer, and Tolstikhin {[}BHM+23{]} use instead infinite
trees in which every node carries \(k\) points, whose largest \(k\) is
the VCL dimension. These trees govern \emph{universal} learning curves,
in which a single distribution is fixed, the learner sees i.i.d. samples
from it, and rates are measured as the sample size grows, separately for
each distribution. Infinite Littlestone trees and infinite VCL trees
separate exponential, linear, and arbitrarily slow universal rates
{[}BHM+21{]}, and the VCL dimension governs the linear term of the
learning curve {[}BHM+23{]}. The epoch dimension counts instead the
points on finite trees of a given depth, and controls worst-case
mistakes when the distribution shifts up to \(n-1\) times.

\textbf{Transductive online learning.} In transductive online learning
{[}BKM97{]}, the adversary fixes the sequence of instances in advance
and shows it to the learner, and then reveals their labels one at a
time. Hanneke, Moran, and Shafer {[}HMS23{]} show that the optimal
number of mistakes on \(T\) instances is \(\Theta(1)\),
\(\Theta(\log T)\), or \(T\), according as the Littlestone dimension is
finite, the VC dimension is finite but the Littlestone dimension is not,
or the VC dimension is infinite. The history game with a single epoch is
a stochastic relative of this model: the adversary commits in advance to
a distribution rather than to a sequence, and does not show it. For
\(T\ge\VC(H)\), Theorem \ref{thm:main} then places the optimum between
\(\frac{1-1/e}{2}\VC(H)\) and
\(\min\big(\Ldim(H),(2+\ln T)\VC(H)\big)\); for thresholds the logarithm
is needed (Proposition \ref{prop:zoom}), as in the middle case of the
trichotomy. Each further epoch lets the adversary commit afresh, and the
epoch dimension measures how far \(n\) commitments move the answer from
the VC dimension toward the Littlestone dimension.

\textbf{Smoothed and constrained adversaries.} Smoothed online learning
lets the adversary choose a new distribution at every round, subject to
a bound on its density with respect to a base measure, and recovers
rates governed by the VC dimension {[}RST11; HRS21; BDGR22{]}. Blanchard
and Kpotufe {[}BK25{]} characterize, more generally, the classes of
distributions from which an adversary may choose at every round while
learning remains possible. Goel, Hanneke, Moran, and Shetty {[}GHMS23{]}
keep a fixed target and an i.i.d. stream, but let an adversary inject
clean-label examples into it; a learner that may abstain on them, and
knows the distribution of the others, makes a number of mistakes
governed by the VC dimension. These models constrain what each
distribution may be, or how the adversary may intervene, at every round.
The history game leaves the distributions unconstrained, but bounds the
number of changes, with i.i.d. examples in between; once
\(n\ge\Ldim(H)\), its lower bound reaches the Littlestone dimension, the
cost of the fully adversarial game.

\textbf{Distribution shift.} When a fixed concept labels the examples, a
change of distribution is a covariate shift: the distribution of the
examples changes, and the conditional distribution of their labels does
not {[}Shi00; QSSL09{]}. Domain adaptation bounds the error on a target
distribution of a hypothesis trained on a source distribution {[}MMR09;
BBC+10{]}, and learning with drifting distributions lets the
distribution move by a bounded amount at every step {[}BL97; MM12{]}.
Concept drift is the dual setting, in which the target changes and the
distribution does not {[}HL94; BBDK00{]}. Hanneke {[}Han21{]} asks, for
a fixed target, for which processes of examples learning is possible at
all; the history game is one such process, chosen by an adversary. In
the history game the shifts are abrupt, adversarial, and few, and the
question is not how well what was learned transfers to a new
distribution, but how many mistakes the shifts can force in all. We are
not aware of earlier work on learning a fixed concept from i.i.d.
examples whose distribution changes adversarially a bounded number of
times.

\textbf{Tracking, adaptive regret, and dynamic regret.} Our learner for
hidden boundaries aggregates, by fixed share {[}HW98{]}, copies of a
base learner started at different rounds. This is the mechanism behind
tracking the best expert and behind adaptive regret, which is low regret
on every interval of time {[}HS09; AKCV16; DGS15; JOWW17{]}. Dynamic
regret compares the learner with a sequence of comparators of bounded
variation {[}Zin03; BGZ15{]}, and switching bandits are the bandit
counterpart of a piecewise stationary environment {[}GM11{]}. In all of
these the benchmark changes with the environment, and a learner pays
afresh after each change. Under one law the benchmark never changes:
what the learner learned in one epoch remains true in the next, and the
adversary can only make it irrelevant. This is why \(n\) epochs cost the
epoch dimension rather than \(n\) times the cost of one: for the cube on
\(d\) points, \(d\) rather than \(nd\).

\section{Conclusion}\label{sec:conclusion}

The epoch dimension \(\ED_n(H)\) counts the questions an adversary can
ask in \(n\) epochs, all at once within an epoch and adaptively across
epochs, with every answer given by a single hypothesis. It interpolates
the VC and Littlestone dimensions, satisfies a simple recursion, is
finite whenever the VC dimension is, and characterizes, up to a factor
\(O(\log T)\), learning a fixed concept from examples whose distribution
shifts at up to \(n-1\) times (Theorem \ref{thm:main}). We now briefly
survey what might come next for this line of work.

\textbf{When the law itself changes.} If every epoch may have a law of
its own, the dimension becomes \(n\VC(H)\) (Remark \ref{rem:perepoch}),
and the same tools show that it characterizes that setting in the same
sense. Fresh labels in every epoch force \(\frac{1-1/e}{2}\,n\VC(H)\)
expected mistakes when \(T\ge n\VC(H)\), as in the proof of Theorem
\ref{thm:lower}, and fixed share over copies of the one-inclusion graph
predictor started at every round makes \(O(n\VC(H)\log T)\). The ratio
of the two dimensions is the price of a law that may change: nothing for
thresholds and intervals, a factor \(n\) for cubes, and a factor tending
to \((d+1)/d\) for half-spaces in \(\mathbb{R}^d\). Moreover, in the
agnostic case, even if the law stays fixed but does not lie in \(H\), a
change of distribution can change which hypothesis of \(H\) fits best.

\textbf{Open problems.} (1) Like Littlestone's standard optimal
algorithm, the layered SOA evaluates dimensions of version spaces. For
which classes can the bound of Theorem \ref{thm:main} be attained
efficiently? (2) For thresholds the logarithmic factor is necessary
(Proposition \ref{prop:zoom}), and by Theorem \ref{thm:main} it can
matter only when \(\Ldim(H)\) is much larger than \(\ED_n(H)\). What
does it multiply? For the product of the thresholds and a cube on \(d\)
points, \(\ED_n=n+d\), but one would expect only the thresholds to pay
the logarithm, for about \(d+n\log(T/n)\) mistakes.

We conclude this paper as we began it: in a world of punctuated
equilibria, our view may shift like water, but nothing learned is ever
unlearned; the epoch dimension measures how much we may yet have to
learn.

\secend

\section*{References}\label{references}
\addcontentsline{toc}{section}{References}

{[}AKCV16{]} D. Adamskiy, W. M. Koolen, A. Chernov, and V. Vovk. A
closer look at adaptive regret. \emph{Journal of Machine Learning
Research}, 2016.

{[}BBC+10{]} S. Ben-David, J. Blitzer, K. Crammer, A. Kulesza, F.
Pereira, and J. W. Vaughan. A theory of learning from different domains.
\emph{Machine Learning}, 2010.

{[}BBDK00{]} P. L. Bartlett, S. Ben-David, and S. R. Kulkarni. Learning
changing concepts by exploiting the structure of change. \emph{Machine
Learning}, 2000.

{[}BDGR22{]} A. Block, Y. Dagan, N. Golowich, and A. Rakhlin. Smoothed
online learning is as easy as statistical learning. \emph{COLT}, 2022.

{[}BGZ15{]} O. Besbes, Y. Gur, and A. Zeevi. Non-stationary stochastic
optimization. \emph{Operations Research}, 2015.

{[}BHM+21{]} O. Bousquet, S. Hanneke, S. Moran, R. van Handel, and A.
Yehudayoff. A theory of universal learning. \emph{STOC}, 2021.

{[}BHM+23{]} O. Bousquet, S. Hanneke, S. Moran, J. Shafer, and I.
Tolstikhin. Fine-grained distribution-dependent learning curves.
\emph{COLT}, 2023.

{[}BK25{]} M. Blanchard and S. Kpotufe. Distributionally-constrained
adversaries in online learning. arXiv:2506.10293, 2025.

{[}BKM97{]} S. Ben-David, E. Kushilevitz, and Y. Mansour. Online
learning versus offline learning. \emph{Machine Learning}, 1997.

{[}BL97{]} R. D. Barve and P. M. Long. On the complexity of learning
from drifting distributions. \emph{Information and Computation}, 1997.

{[}DGS15{]} A. Daniely, A. Gonen, and S. Shalev-Shwartz. Strongly
adaptive online learning. \emph{ICML}, 2015.

{[}EG72{]} N. Eldredge and S. J. Gould. Punctuated equilibria: an
alternative to phyletic gradualism. In T. J. M. Schopf (ed.),
\emph{Models in Paleobiology}, Freeman, Cooper and Co., 1972.

{[}GE77{]} S. J. Gould and N. Eldredge. Punctuated equilibria: the tempo
and mode of evolution reconsidered. \emph{Paleobiology}, 1977.

{[}GHMS23{]} S. Goel, S. Hanneke, S. Moran, and A. Shetty. Adversarial
resilience in sequential prediction via abstention. \emph{NeurIPS},
2023.

{[}GM11{]} A. Garivier and E. Moulines. On upper-confidence bound
policies for switching bandit problems. \emph{ALT}, 2011.

{[}Han21{]} S. Hanneke. Learning whenever learning is possible:
universal learning under general stochastic processes. \emph{Journal of
Machine Learning Research}, 2021.

{[}HL94{]} D. P. Helmbold and P. M. Long. Tracking drifting concepts by
minimizing disagreements. \emph{Machine Learning}, 1994.

{[}HLW94{]} D. Haussler, N. Littlestone, and M. K. Warmuth. Predicting
\(\{0,1\}\)-functions on randomly drawn points. \emph{Information and
Computation}, 1994.

{[}HMS23{]} S. Hanneke, S. Moran, and J. Shafer. A trichotomy for
transductive online learning. \emph{NeurIPS}, 2023.

{[}HRS21{]} N. Haghtalab, T. Roughgarden, and A. Shetty. Smoothed
analysis with adaptive adversaries. \emph{FOCS}, 2021.

{[}HS09{]} E. Hazan and C. Seshadhri. Efficient learning algorithms for
changing environments. \emph{ICML}, 2009.

{[}HW98{]} M. Herbster and M. K. Warmuth. Tracking the best expert.
\emph{Machine Learning}, 1998.

{[}JOWW17{]} K.-S. Jun, F. Orabona, S. Wright, and R. Willett. Improved
strongly adaptive online learning using coin betting. \emph{AISTATS},
2017.

{[}Lit88{]} N. Littlestone. Learning quickly when irrelevant attributes
abound: a new linear-threshold algorithm. \emph{Machine Learning}, 1988.

{[}MM12{]} M. Mohri and A. Muñoz Medina. New analysis and algorithm for
learning with drifting distributions. \emph{ALT}, 2012.

{[}MMR09{]} Y. Mansour, M. Mohri, and A. Rostamizadeh. Domain
adaptation: learning bounds and algorithms. \emph{COLT}, 2009.

{[}QSSL09{]} J. Quiñonero-Candela, M. Sugiyama, A. Schwaighofer, and N.
D. Lawrence, eds.~\emph{Dataset Shift in Machine Learning}. MIT Press,
2009.

{[}Rad21{]} J. Radon. Mengen konvexer Körper, die einen gemeinsamen
Punkt enthalten. \emph{Mathematische Annalen}, 1921.

{[}RST11{]} A. Rakhlin, K. Sridharan, and A. Tewari. Online learning:
stochastic, constrained, and smoothed adversaries. \emph{NeurIPS}, 2011.

{[}Shi00{]} H. Shimodaira. Improving predictive inference under
covariate shift by weighting the log-likelihood function. \emph{Journal
of Statistical Planning and Inference}, 2000.

{[}VC71{]} V. N. Vapnik and A. Ya. Chervonenkis. On the uniform
convergence of relative frequencies of events to their probabilities.
\emph{Theory of Probability and its Applications}, 1971.

{[}Zin03{]} M. Zinkevich. Online convex programming and generalized
infinitesimal gradient ascent. \emph{ICML}, 2003.

\end{document}